\pdfoutput=1
\documentclass[11pt,a4paper]{article}

\usepackage[utf8]{inputenc}
\usepackage[T1]{fontenc}
\usepackage{lmodern}
\input{t1lmr.fd}
\DeclareFontShape{T1}{lmr}{bx}{sc}{<->ssub*lmr/bx/n}{}
\usepackage[margin=1in]{geometry}
\usepackage{microtype}
\AtBeginDocument{\renewcommand{\_}{\textunderscore\allowbreak}}
\begingroup
\catcode`\-=\active
\catcode`\/=\active
\gdef\weaveTTbreaks{%
  \catcode`\-=\active \catcode`\/=\active
  \def-{\char`\-\allowbreak}%
  \def/{\char`\/\allowbreak}%
}
\endgroup
\NewCommandCopy{\weaveOrigTexttt}{\texttt}
\DeclareRobustCommand{\texttt}{\begingroup\weaveTTbreaks\weaveTextttArg}
\newcommand{\weaveTextttArg}[1]{\weaveOrigTexttt{\scantokens{#1\noexpand}}\endgroup}
\usepackage{amsmath,amssymb,amsfonts,amsthm}
\usepackage{booktabs}
\usepackage{tabularx}
\usepackage{xcolor}
\usepackage{graphicx}
\usepackage{pgfplots}
\pgfplotsset{compat=1.18}
\usepackage{listings}
\usepackage{upquote}

\usepackage{tikz}
\usetikzlibrary{
    arrows.meta,
    positioning,
    fit,
    calc,
    shapes.symbols,
    backgrounds
}

\definecolor{weaveLink}{RGB}{15, 55, 115}

\lstdefinestyle{weaveListing}{
    backgroundcolor=\color{black!3},
    basicstyle=\ttfamily\small,
    breaklines=true,
    captionpos=b,
    frame=single,
    rulecolor=\color{black!25},
    numbers=left,
    numberstyle=\tiny\color{black!55},
    keywordstyle=\color{black}\bfseries,
    commentstyle=\color{black!65}\itshape,
    stringstyle=\color{black!85},
    showstringspaces=false,
    tabsize=2
}
\usepackage[
    colorlinks=true,
    linkcolor=weaveLink,
    citecolor=weaveLink,
    urlcolor=weaveLink,
    pdfauthor={Martin Lukasiewycz},
    pdftitle={WeaveMX: Zero-Code Orchestration of Structured Multi-Agent Systems for Asset Generation},
    pdfsubject={cs.AI; cs.MA; cs.SE},
    pdfkeywords={Multi-Agent Systems, Declarative Orchestration, Agent Harnesses, Model Context Protocol, Provenance Graphs, Human-in-the-Loop}
]{hyperref}

\newcommand{\weave}{\textsc{WeaveMX}}
\newcommand{\mcpweave}{\texttt{weave}}

\title{\textbf{\weave{}: Zero-Code Orchestration of Structured Multi-Agent Systems for Asset Generation}}

\author{
    \textbf{Martin Lukasiewycz} \\
    \texttt{martin.lukasiewycz@axiomfab.org}
}

\date{}

\begin{document}

\maketitle

\begin{abstract}
Single-prompt Large Language Model (LLM) inference works well for self-contained queries, but generating complex, multi-part digital assets requires coordinated multi-agent systems (MAS). Existing tools pull in two directions. Code-centric state graphs demand heavy engineering and costly optimization. Open-ended harnesses and single-agent loops avoid that cost, yet suffer from inconsistent runs, lossy handoffs, and poor visibility into progress. As a remedy, this paper presents \weave{} (Multi-agent eXecution), a declarative orchestration layer for agent harnesses whose main contribution is the zero-code definition of structured MAS in plain Markdown. Authors define orchestrators and isolated workers in partially predefined Markdown files that pair natural-language instructions with file and parameter ports, written by hand or from a brief. Dedicated skills guide agents through a predictable protocol, while programmatic Model Context Protocol (MCP) tools isolate worker directories, check port contracts, and track lifecycles in on-disk Markdown manifests. Declared port signatures induce cyclic workflow graphs that unwind at runtime into acyclic provenance graphs under concurrent execution and subset I/O, grounding static auditing and runtime checks. Because the LLM orchestrator reasons over files and resumable sessions rather than a compiled state machine, \weave{} supports both batch runs and interactive human--AI workflows. Creators inspect live assets in a zero-token filesystem UI, steer at review gates, and make cross-stage surgical edits in natural language without pre-wired edges or checkpoint replay. \weave{} supports Google Antigravity (\texttt{agy}) app and IDE, VS Code, and Cursor out of the box, with setup notes for Claude Code and Windsurf. Case studies in \texttt{agy} evaluate \weave{} across rapid pipeline synthesis and production-scale interactive asset generation.
\end{abstract}

\section{Introduction}
\label{sec:intro}

Prompting a single Large Language Model (LLM) to produce a complex creative or analytical deliverable in one pass forces competing requirements into a single context window: tone, structure, factual accuracy, multimodal planning, and self-critique. While single-prompt inference handles short, self-contained tasks well, generating rich, multi-part digital assets quickly exposes the limits of monolithic generation.

\subsection{Why Go Beyond a Single LLM?}
\label{sec:intro_why_mas}

Splitting work across a coordinated Multi-Agent System (MAS)~\cite{park2023generative,wu2023autogen,hong2023metagpt} addresses practical bottlenecks that arise when forcing an entire workflow into one model call. As conversation history and generated text accumulate inside a single context window, models begin dropping negative constraints, skipping sections, or hitting output token limits mid-task. Breaking the job into smaller units lets an orchestrator run isolated agents sequentially or in parallel (for instance, ten chapter writers working concurrently), each starting with a clean context window. Different stages also call for different compute profiles: open-ended research benefits from stronger reasoning models with web search, drafting favors faster general-purpose models, and illustration requires image generation tools. Assigning each specialist only the model tier and toolset its role needs avoids bloated tool schemas and unnecessary inference spend.

Task decomposition further improves factual grounding and inspectability. When a single prompt must be both imaginative and strictly factual, accuracy often suffers: in the \textsc{KidsWiki} example (Section~\ref{sec:formal_running_example}), prompting a frontier LLM in one turn to explain ``Rockets'' to a 4-year-old yields engaging but false prose (\emph{``A rocket is a super-strong space boat that flies all the way to the stars!''}), whereas pairing the \textbf{Writer} with an independent \textbf{Fact Checker} catches and repairs the claim in a targeted revision loop. Persisted intermediate artifacts (\texttt{01\_draft.md}, \texttt{fact\_audit.md}, \texttt{02\_illustration.png}) likewise give human operators concrete checkpoints to inspect and steer the work rather than regenerating everything from scratch when an early assumption goes wrong. Added reliability and control come with a latency trade-off: coordinating isolated workspaces, checking file contracts, running verification loops, and pausing at review gates often takes substantially longer than a single prompt.

\subsection{Iterative Single-Agent Harness Loops vs.\ Multi-Agent Systems}
\label{sec:intro_ralph_loops}

A popular alternative to multi-agent orchestration is the \emph{Iterative Single-Agent Harness Loop}, often called a ``Ralph loop'' by practitioners, where a single agent with shell and filesystem access is re-invoked in an outer loop against a persistent task list on disk until every item is checked off. Externalizing state to disk and resetting context between passes lets a single agent work past single-turn token limits without multi-agent coordination logic.

Even so, single-agent loops fall short of structured multi-agent systems in multiple ways.

\paragraph{Unpredictable and Unrepeatable Cross-Run Quality.}
Because the looping agent improvises its task decomposition and tool sequence on every run, one successful pass does not guarantee repeatable quality: running the same loop on a different input gives no assurance of following a comparable workflow or achieving similar output quality.

\paragraph{Reliance on Automated Checks and Poor Fit for Human Interaction.}
Autonomous outer loops work best when completion can be checked mechanically (such as passing unit tests or compiler checks), which rarely holds for creative asset generation. Moreover, unattended checklist loops are not designed around structured human review gates or interactive mid-run steering.

\paragraph{Correlated Blind Spots.}
Across fresh context passes, a single agent running under one persona and full workspace visibility tends to share its own blind spots, often signing off on subtle hallucinations or quietly relaxing rubric criteria to mark tasks done. Reliable verification requires an independent auditor whose prompt, tools, and input view are separated from the generator's.

\paragraph{Monolithic Tool and Model Assignment.}
A single looping agent needs the union of all tools and permissions used anywhere in the workflow and typically runs every iteration on one frontier model, ruling out per-step model selection or parallel execution across independent subtasks.

\paragraph{Unrestricted Workspace Access.}
Because the looping agent can read and write anywhere in the project tree, a mistaken edit in a later pass can overwrite upstream instructions, alter validated outputs, or pollute shared files without boundary checks.

\paragraph{Compliance and Least-Privilege Governance.}
Domains such as finance, healthcare, legal publishing, and safety-critical engineering require least-privilege access control, separation of duties, and clear audit trails. Constraining each agent through an explicit input/output signature ensures that external-facing agents cannot read confidential files, no agent writes outside of its working directory, and every deliverable passes through a separate verification step recorded on disk.

\subsection{The Multi-Agent Landscape and the Limits of Dynamic Flexibility}
\label{sec:intro_landscape}

Current multi-agent tools largely fall into two camps, neither of which fits rapid iteration and interactive digital asset production well.

\paragraph{1. Code-Centric Graph and Conversational Frameworks.}
Libraries such as LangGraph~\cite{langchain2024langgraph}, AutoGen~\cite{wu2023autogen}, and MetaGPT~\cite{hong2023metagpt} provide programmatic abstractions for multi-agent workflows. Code-based state graphs offer static typing, persistent state checkpointing, and headless execution suited to production backend services, but creating or changing a workflow means writing Python or TypeScript glue code, managing dependencies, and wiring up state serialization. Even when checkpointing supports rewinding to an earlier state, resuming a compiled graph replays downstream transitions across serialized state dictionaries unless custom skip logic and dirty-bit guards were pre-coded in advance. For interactive human--AI creative processes (where creators routinely step back several stages to make unanticipated stylistic or structural tweaks), that engineering rigidity gets in the way.

\paragraph{2. Open-Ended Dynamic Agent Harnesses.}
General-purpose agent environments such as Google Antigravity (\texttt{agy})~\cite{google2024antigravity}, Claude Code, Cursor, and OpenHands~\cite{wang2024openhands} prioritize open-ended flexibility for tasks like interactive debugging, codebase exploration, and ad-hoc research. In \texttt{agy}, for instance, an Editor view and an Agent Manager can spawn subagents on demand, complemented by a built-in \emph{Swarm} mode (where parallel workers coordinate over a shared message board) and an \emph{Orchestrator} that switches dynamically between model tiers at runtime.

Ad-hoc subagent dispatch, message-board swarms, and heuristic model switching work well for exploratory coding, yet break down in predictable ways when a workflow requires repeatable structure across runs. Without explicit stage contracts, repeated invocations drift across different folder layouts, skip intermediate deliverables, or bypass verification steps. When an orchestrator passes context by summarizing prior turns in chat, detailed constraints, citations, and formatting rules are gradually compressed away. Left unconstrained, subagents also spend turns scanning parent directories (\texttt{ls}, \texttt{dir}), reading unrelated repository files for orientation, or spawning fresh instances during revisions instead of continuing existing sessions, while operators lack on-disk manifests to tell which stage is running or where a stalled run stopped without digging through raw transcripts.

\subsection{How \weave{} is Different: Predictability Without Code}
\label{sec:intro_weavemas}

This paper presents \weave{} (Multi-agent eXecution), a declarative orchestration layer for agent harnesses whose primary contribution is the zero-code definition of structured multi-agent systems in plain Markdown. Running across multiple MCP-capable IDE and CLI harnesses (with Google Antigravity (\texttt{agy}) as the primary reference environment in this paper), \weave{} achieves predictable execution without orchestration code through a three-part causal design. First, \emph{partially predefined declarative Markdown files} define the orchestrator (\texttt{orchestrator.md}, \texttt{orchestrator/stage*.md}) and specialist worker agents (\texttt{.agent/agents/*.md}), combining structured headers for stages, roles, file/parameter ports, tasks (\texttt{\# Role Specifics}, \texttt{\# Tasks}, \texttt{\# Lifecycle}), and review gates with natural-language instructions (written by hand or synthesized from a short brief via \texttt{weave-authoring}). Second, \emph{orchestrator and agent execution skills} (\texttt{weave-orchestrator} and \texttt{weave-agent}) govern how the host harness interprets those Markdown files, prescribing repeatable operating protocols for workspace provisioning, subagent dispatch, directory discipline, and multi-turn handoffs. Third, \emph{programmatic Weave MCP tools}~\cite{anthropic2024mcp} (\mcpweave{}) enforce those protocols at runtime, sandboxing each worker inside its own Current Working Directory (CWD), validating declared input/output port contracts, and managing orchestrator and worker lifecycles in on-disk \texttt{weave.md} manifests. Because \texttt{weave.md} externalizes lifecycle state to disk, an out-of-band Node.js UI polls the local filesystem to display stage progress, per-agent status cards, elapsed runtimes, and single-click artifact previews with zero LLM token overhead.

For interactive creative workflows, \weave{} pairs human review gates with an LLM orchestrator reasoning over discrete files rather than compiled state transitions. Stage gates pause execution at verified milestone deliverables (such as drafted prose and embedded image prompts) before triggering costly downstream synthesis, while persisted intermediate files and resumable worker sessions keep retries local to individual stages. At any review gate or after a run finishes, a creator can also request cross-stage surgical edits in plain language (such as revising one upstream analogy or regenerating a single illustration) without altering the reusable pipeline specification: the orchestrator traces declared file dependencies, wakes the relevant worker session in place, and updates only affected downstream steps.

\subsection{Contributions of the Paper}
\label{sec:intro_contributions}

This paper makes four main contributions, led by the zero-code specification of structured multi-agent workflows:
\begin{enumerate}
    \item \textbf{Zero-Code Declarative Multi-Agent Specification (Section~\ref{sec:implementation}):} Introduces a partially predefined Markdown format for defining orchestrators, multi-stage plans, and isolated worker agents without orchestration code, supporting both direct hand-authoring and one-shot pipeline synthesis with contract-preserving revision via \texttt{weave-authoring}.
    \item \textbf{Predictable Execution via Skills, MCP Enforcement, and \texttt{weave.md} Lifecycles (Section~\ref{sec:implementation}):} Establishes a causal runtime stack in which orchestrator and worker skills (\texttt{weave-orchestrator}, \texttt{weave-agent}) prescribe repeatable two-phase provisioning and dispatch protocols, programmatic MCP tools (\mcpweave{}) enforce per-agent CWD sandboxing, scoped port bindings, and on-disk \texttt{weave.md} lifecycle state, and a filesystem-polling UI provides zero-token observability.
    \item \textbf{Formal Port-Graph Semantics and Two-Tier Consistency Auditing (Sections~\ref{sec:formal_model} and~\ref{sec:verification}):} Formalizes Markdown file and parameter port signatures as cyclic tripartite graphs that unwind at runtime into acyclic provenance DAGs under concurrent execution and selective subset I/O, grounding authoring-time data-flow checks (Port Contract Isomorphism, Causal Input Availability, and Monotonic State Evolution) alongside runtime MCP enforcement and future pipeline optimization (Section~\ref{sec:conclusion}).
    \item \textbf{End-to-End Case Studies (Section~\ref{sec:case_studies}):} Evaluates \weave{} across two pipelines: an introductory 2-stage workflow (\textsc{KidsWiki}, illustrating one-shot synthesis, parallel fact and age verification loops, and human-gated illustration) and a 9-stage, 17-role production workflow (\textsc{TheTuneMachine}, examining scale, UI observability, per-role token telemetry across model tiers, and cross-stage surgical edits).
\end{enumerate}

The \weave{} plugin, skills, and \mcpweave{} MCP server are publicly available under the MIT license at \url{https://github.com/axiomfab/weavemx}.

The rest of the paper is organized as follows. Section~\ref{sec:related_work} situates \weave{} within related work. Section~\ref{sec:formal_model} introduces the formal system model, Section~\ref{sec:implementation} details the implementation, and Section~\ref{sec:verification} presents the data-flow auditing and runtime enforcement mechanisms. Section~\ref{sec:case_studies} presents the end-to-end case studies, and Section~\ref{sec:conclusion} outlines future directions and concluding remarks.

\section{Related Work}
\label{sec:related_work}

\weave{} connects four lines of prior work: multi-agent LLM frameworks, open-ended agent harnesses and single-agent loops, deterministic scientific workflow engines, and capability sandboxing with tool protocols. Table~\ref{tab:related_work_comparison} compares \weave{} against representative systems across these areas.

\subsection{Code-Centric Multi-Agent Frameworks and Conversational Orchestration}
\label{sec:rw_mas_frameworks}

Early multi-agent LLM work simulated believable social interaction across autonomous personas~\cite{park2023generative} and showed that splitting complex tasks across distinct agent roles improves reasoning diversity and output quality over single-prompt inference~\cite{li2023camel,wu2023autogen,chen2024agentverse}. Conversational frameworks such as CAMEL~\cite{li2023camel}, AutoGen~\cite{wu2023autogen}, AgentVerse~\cite{chen2024agentverse}, and DyLAN~\cite{liu2024dylan} coordinate agents mainly through multi-turn dialogue. Free-form dialogue makes initial experimentation easy, but passing workflow state through growing chat transcripts leads to context bloat and lossy summarization: across turns, formatting rules, negative constraints, and source provenance gradually erode unless anchored to external files.

To add stronger structure, role- and Standard Operating Procedure (SOP) frameworks such as MetaGPT~\cite{hong2023metagpt}, ChatDev~\cite{qian2024chatdev}, and CrewAI~\cite{crewai2024} assign agents to specialized roles or waterfall software-engineering phases (as in MetaGPT and ChatDev) that exchange structured intermediate outputs. In parallel, graph- and compiler-oriented frameworks including LangGraph~\cite{langchain2024langgraph}, Graph of Thoughts (GoT)~\cite{besta2024got}, GPTSwarm~\cite{zhuge2024gptswarm}, and DSPy~\cite{khattab2024dspy} model multi-step LLM programs as explicit state graphs or optimizable pipelines.

While expressive, all of these systems are \emph{code-centric}: defining topologies, conditional cycles, and state schemas requires writing Python or TypeScript orchestration code, maintaining package environments, and handling state serialization. Optimization frameworks such as DSPy~\cite{khattab2024dspy} and GPTSwarm~\cite{zhuge2024gptswarm} further assume multi-example training runs whose API token costs rarely pay off for interactive, small-batch digital asset creation. By contrast, \weave{} eliminates orchestration code altogether: pipelines are written in plain Markdown, generated from a concise brief, and audited before execution while preserving formal graph semantics ($G \to \hat{G}$).

\subsection{Open-Ended Agent Harnesses and Iterative Single-Agent Loops}
\label{sec:rw_single_agent_loops}

A second line of work equips an iterative reasoning-and-acting loop, building on ReAct~\cite{yao2023react}, Reflexion~\cite{shinn2023reflexion}, Self-Refine~\cite{madaan2023selfrefine}, and Voyager~\cite{wang2023voyager}, with direct shell and filesystem tools. Developer harnesses such as SWE-agent~\cite{yang2024sweagent}, OpenHands~\cite{wang2024openhands}, and Google Antigravity (\texttt{agy})~\cite{google2024antigravity} expose file editing and command execution inside an interactive workspace, with platforms such as OpenHands and \texttt{agy} further adding subagent dispatch. Practitioners often run these harnesses as \emph{iterative single-agent loops} (``Ralph loops''), repeatedly invoking one agent against a persistent checklist on disk until every item is finished.

Although outer-loop repetition avoids multi-agent setup, prior studies show clear limits to single-agent self-correction. Huang et al.~\cite{huang2024selfcorrect}, for example, show that without independent external feedback, LLMs struggle to self-correct reasoning mistakes and can even degrade accuracy across self-critique rounds. When generator and critic share the same system persona, tools, and full-workspace view, subtle factual slips (such as the interstellar rocket claim in the \textsc{KidsWiki} example) easily pass review. Unrestricted workspace access also lets the agent wander across unrelated directories or accidentally overwrite upstream instructions. \weave{} builds directly on the subagent and MCP primitives of modern developer harnesses (such as Google Antigravity~\cite{google2024antigravity}), bounding execution through separate specialist roles, per-agent CWD sandboxing, and scoped file resolution.

\begin{table*}[t]
\centering
\caption{Taxonomy and Architectural Comparison of Multi-Agent, Single-Agent, and Workflow Paradigms.}
\label{tab:related_work_comparison}
\footnotesize
\setlength{\tabcolsep}{4pt}
\renewcommand{\arraystretch}{1.22}
\begin{tabularx}{\textwidth}{@{}>{\raggedright\arraybackslash}p{2.45cm} >{\raggedright\arraybackslash}p{2.15cm} >{\raggedright\arraybackslash}X >{\raggedright\arraybackslash}X >{\raggedright\arraybackslash}p{1.85cm} >{\raggedright\arraybackslash}X >{\raggedright\arraybackslash}p{2.35cm}@{}}
\toprule
\textbf{System / Paradigm} &
\textbf{Spec.\ Modality} &
\textbf{State Handoff} &
\textbf{Topology \& Cycles} &
\textbf{Pre-Run Check} &
\textbf{I/O Sandboxing} &
\textbf{Observability Mechanism} \\
\midrule
\textbf{Conversational MAS} \newline {\scriptsize (AutoGen~\cite{wu2023autogen}, CAMEL~\cite{li2023camel})} &
Python API + Prompts &
Chat history / dialogue turns &
Dynamic chat graph (cyclic) &
None (runtime only) &
Shared process / optional Docker &
In-band chat transcripts \& callback loggers \\
\addlinespace[3pt]
\textbf{SOP Role Frameworks} \newline {\scriptsize (MetaGPT~\cite{hong2023metagpt}, ChatDev~\cite{qian2024chatdev})} &
Python classes + SOP schemas &
Structured msgs \& shared repo &
Fixed waterfall / phase loops &
Schema type checks &
Shared workspace directory &
Console streams \& web replay logs \\
\addlinespace[3pt]
\textbf{Code State Graphs} \newline {\scriptsize (LangGraph~\cite{langchain2024langgraph}, DSPy~\cite{khattab2024dspy})} &
Imperative Python / TS &
In-memory state / typed fields &
Explicit state / compute graph &
Host-language types &
Application-defined scope &
Cloud telemetry traces~\cite{dong2024agentops} (e.g., LangSmith) \\
\addlinespace[3pt]
\textbf{Iterative Agent Loops} \newline {\scriptsize (SWE-agent~\cite{yang2024sweagent}, ``Ralph'' loops)} &
Single prompt + outer bash loop &
Global mutable filesystem &
Unbounded single-agent loop &
None &
Unrestricted repo read/write &
Raw terminal \& trajectory JSONL logs \\
\addlinespace[3pt]
\textbf{Scientific Workflows} \newline {\scriptsize (Nextflow~\cite{di2017nextflow}, Snakemake~\cite{koster2012snakemake})} &
Custom DSL (Groovy / Python) &
File channels / target rules &
Acyclic job DAG &
DAG cycle \& file rule check &
Task workdirs (Nextflow) / opt-in shadow &
CLI progress \& execution reports \\
\midrule
\textbf{\weave{} (This work)} &
\textbf{Plain Markdown (zero code)} &
\textbf{Physical files ($\mathcal{F}$) \& params ($\mathcal{P}$)} &
\textbf{Cyclic $G \to$ Unwound DAG $\hat{G}$} &
\textbf{Port Isomorphism \& Causal Flow} &
\textbf{Per-agent CWD + Scoped MCP Ports} &
\textbf{Async filesystem UI} \\
\bottomrule
\end{tabularx}
\end{table*}

\subsection{Deterministic Dataflow, Scientific Workflows, and File-Grounded State}
\label{sec:rw_workflows}

The formal model draws directly on classical concurrency models and scientific workflow managers. In Kahn Process Networks (KPN)~\cite{kahn1974semantics}, concurrent processes communicate strictly over unbounded FIFO channels with blocking reads, guaranteeing determinate input--output histories regardless of scheduling order, whereas Petri Nets~\cite{murata1989petri} formalize concurrent token-driven synchronization and firing preconditions across place--transition graphs. Classical Blackboard systems~\cite{nii1986blackboard} likewise decouple specialist modules by coordinating through a shared structured store rather than direct peer-to-peer messages.

In computational science, file-based workflow engines such as Snakemake~\cite{koster2012snakemake} and Nextflow~\cite{di2017nextflow} achieve reproducibility and local fault recovery by modeling steps as explicit file-to-file transformations, executing tasks either in isolated per-process work directories by default (Nextflow) or in a shared workspace with opt-in shadow directories and containers (Snakemake). Even where dynamic checkpoints or channels defer graph construction, classical scientific workflow engines still require an \emph{acyclic} dependency graph (DAG): they cannot naturally express cyclic generator--verifier revision loops, iteration-dependent subset I/O over cyclic ports, or stateful multi-turn LLM sessions that pause in \texttt{waiting/*} states and resume on incoming messages. \weave{} bridges these two worlds by expressing the static specification $G = (V, E)$ as a cyclic tripartite graph over agents $\mathcal{A}$, files $\mathcal{F}$, and parameters $\mathcal{P}$ that unwinds dynamically at runtime into an acyclic provenance DAG $\hat{G} = (\hat{V}, \hat{E})$ (Propositions~1 and~2), while still supporting authoring-time causal data-flow checks (Section~\ref{sec:arch_verification}).

\subsection{Tool Protocols, Capability Sandboxing, and Out-of-Band Observability}
\label{sec:rw_mcp_security_obs}

Connecting LLMs to external tools has moved from model-specific fine-tuning in Toolformer~\cite{schick2023toolformer} and Gorilla~\cite{patil2024gorilla} to standardized protocols, most notably the Model Context Protocol (MCP)~\cite{anthropic2024mcp}. MCP standardizes how hosts expose tools and resources to models, but leaves workflow ordering and per-agent filesystem isolation up to the application.

From a security standpoint, giving autonomous agents broad tool and filesystem permissions violates least privilege~\cite{saltzer1975protection} and information-flow control~\cite{sabelfeld2003language}. Recent work such as IsolateGPT~\cite{wu2024isolategpt} shows how shared, unisolated LLM environments invite cross-agent state contamination, prompt injection escalation, and unintended data exposure. \weave{} enforces least-privilege boundaries in both specification and execution: an agent's static signature $a = \langle F_{\text{in}}, F_{\text{out}}, P_{\text{in}}, P_{\text{out}} \rangle$ fixes what it may read and write, while the \mcpweave{} MCP server checks inputs before execution (\texttt{weave\_verify}), resolves paths via explicit scope prefixes and port aliases, and validates outputs on completion (\texttt{weave\_update}) inside isolated working directories.

Finally, most agent observability tools~\cite{dong2024agentops} depend on SDK callbacks, cloud tracing services, or supervisor LLMs that summarize peer progress, adding code dependencies or extra token spend. Recording lifecycle state directly in on-disk \texttt{weave.md} manifests lets \weave{} drive a live stage stepper, runtime counters, and artifact previews from a separate UI through simple filesystem polling.

\section{Formal System Model}
\label{sec:formal_model}

To ground zero-code Markdown specifications in rigorous data-flow semantics, this section formalizes \weave{} workflows in three steps: (1)~the passive state artifacts and worker agents defined by their permissible input/output interfaces, (2)~the induced static workflow graph $G = (V, E)$, which allows cycles for iterative refinement, and (3)~the runtime unwinding of $G$ into a time-indexed Directed Acyclic Graph (DAG) $\hat{G} = (\hat{V}, \hat{E})$ under concurrent execution of independent agents and selective subset I/O. Throughout this section, each concept is illustrated on the \textsc{KidsWiki} pipeline.

\subsection{Running Example: The \textsc{KidsWiki} Asset Pipeline}
\label{sec:formal_running_example}

Consider generating factually verified, age-calibrated, illustrated encyclopedia articles for children (\textsc{KidsWiki}). Writing for young readers requires balancing engaging simplicity against scientific accuracy. As seen in Section~\ref{sec:intro}, a single LLM pass explaining ``Rockets'' to a 4-year-old readily invents claims such as rockets flying ``all the way to the stars.''

To separate storytelling from verification without custom orchestration code, \textsc{KidsWiki} coordinates four specialist roles across two stages. In Stage~1 (\emph{Editorial}), a \textbf{Writer} ($a_{\text{writer}}$) takes a target topic and reader age, drafts the article, and embeds an in-place Markdown image tag whose alt-text carries a self-contained visual prompt (\texttt{![Detailed prompt](./02\_illustration.png)}) followed by an italicized caption. Two independent reviewers, a \textbf{Fact Checker} ($a_{\text{fact}}$) and an \textbf{Age Checker} ($a_{\text{age}}$), then audit the draft concurrently for scientific accuracy and age-appropriate readability, writing findings to \texttt{fact\_audit.md} and \texttt{age\_audit.md} and triggering an in-place revision round with $a_{\text{writer}}$ whenever issues arise. Once a human operator approves the verified text at the Stage~1 review gate, Stage~2 (\emph{Publishing}) dispatches an \textbf{Illustrator} ($a_{\text{illustrator}}$) that extracts the alt-text prompt, generates \texttt{02\_illustration.png}, and completes the illustrated article without modifying the prose.

\subsection{Agents and Permissible I/O Signatures}
\label{sec:formal_agents}

Let $\mathcal{F}$ be the set of \emph{files} (persistent unstructured or structured artifacts on disk) and $\mathcal{P}$ be the set of \emph{parameters} (scalar configuration values, prompt arguments, or control flags), with $\mathcal{F} \cap \mathcal{P} = \emptyset$.

Let $\mathcal{A}$ be a finite set of autonomous \emph{worker agents} disjoint from artifacts, $\mathcal{A} \cap (\mathcal{F} \cup \mathcal{P}) = \emptyset$. Each agent $a \in \mathcal{A}$ is characterized by a static 4-tuple defining its permissible input/output interface:
\begin{equation}
    a = \langle F_{\text{in}}(a), \; F_{\text{out}}(a), \; P_{\text{in}}(a), \; P_{\text{out}}(a) \rangle
\end{equation}
Here $F_{\text{in}}(a), F_{\text{out}}(a) \subseteq \mathcal{F}$ denote the maximal sets of files that $a$ may read and write, and $P_{\text{in}}(a), P_{\text{out}}(a) \subseteq \mathcal{P}$ denote the maximal sets of parameters that $a$ may read and write. Any of the four sets may be empty ($\emptyset$). Each element $a \in \mathcal{A}$ represents a distinct stateful agent instance with its own workspace and I/O bindings, even when multiple instances share the same underlying role instructions (for example, a stage dispatching $15$ segment fact-checkers instantiates $15$ distinct elements $a_{\text{fact},1}, \dots, a_{\text{fact},15} \in \mathcal{A}$).

In \textsc{KidsWiki}, the parameter set is $\mathcal{P} = \{\texttt{topic}, \texttt{target\_age}\}$ and the file set is
\begin{equation}
    \mathcal{F} = \{\texttt{01\_draft.md}, \; \texttt{fact\_audit.md}, \; \texttt{age\_audit.md}, \; \texttt{02\_illustration.png}\}
\end{equation}
with static agent signatures:
\begin{align}
    a_{\text{writer}} &= \langle \{\texttt{fact\_audit.md}, \texttt{age\_audit.md}\}, \; \{\texttt{01\_draft.md}\}, \; \{\texttt{topic}, \texttt{target\_age}\}, \; \emptyset \rangle \\
    a_{\text{fact}} &= \langle \{\texttt{01\_draft.md}\}, \; \{\texttt{fact\_audit.md}\}, \; \{\texttt{topic}\}, \; \emptyset \rangle \\
    a_{\text{age}} &= \langle \{\texttt{01\_draft.md}\}, \; \{\texttt{age\_audit.md}\}, \; \{\texttt{target\_age}\}, \; \emptyset \rangle \\
    a_{\text{illustrator}} &= \langle \{\texttt{01\_draft.md}\}, \; \{\texttt{02\_illustration.png}\}, \; \emptyset, \; \emptyset \rangle
\end{align}
These signatures express logical inter-role dataflow. At the physical runtime layer (Sections~\ref{sec:implementation}--\ref{sec:verification}), local working files such as \texttt{draft.md} map onto \texttt{01\_draft.md} via port bindings, while in-stage revision feedback from \texttt{fact\_audit.md} and \texttt{age\_audit.md} is forwarded to the active writer session via message wakeup rather than bound as an initial Turn-1 file input.

\subsection{The Induced Static Workflow Graph \texorpdfstring{$G = (V, E)$}{G = (V, E)}}
\label{sec:formal_static_graph}

A set of worker agents $\mathcal{A}$ induces a static directed tripartite graph $G = (V, E)$ capturing all permissible dataflow pathways. Its vertex set $V$ partitions into three disjoint sets:
\begin{equation}
    V = \mathcal{A} \cup \mathcal{F} \cup \mathcal{P}
\end{equation}
where $\mathcal{F} = \bigcup_{a \in \mathcal{A}} (F_{\text{in}}(a) \cup F_{\text{out}}(a))$ and $\mathcal{P} = \bigcup_{a \in \mathcal{A}} (P_{\text{in}}(a) \cup P_{\text{out}}(a))$.

\begin{figure}[t]
\centering
\begin{tikzpicture}[
    font=\normalsize,
    paramNode/.style={
        signal,
        signal to=east and west,
        draw=black,
        line width=1.1pt,
        fill=white,
        inner sep=4.5pt,
        align=center
    },
    agentNode/.style={
        rectangle,
        rounded corners=6pt,
        draw=black,
        line width=1.3pt,
        fill=white,
        minimum width=3.0cm,
        minimum height=1.05cm,
        align=center,
        inner sep=4pt
    },
    fileNode/.style={
        rectangle,
        draw=black,
        line width=1.1pt,
        fill=white,
        minimum width=3.0cm,
        minimum height=0.95cm,
        align=center,
        inner sep=4pt
    },
    gateNode/.style={
        rectangle,
        draw=black,
        dashed,
        line width=1.2pt,
        fill=black!5,
        minimum width=4.0cm,
        minimum height=0.95cm,
        align=center,
        inner sep=4pt
    },
    arrow/.style={
        -{Stealth[length=6pt, width=4.5pt]},
        line width=1.1pt,
        draw=black
    },
    cycleArrow/.style={
        -{Stealth[length=6.5pt, width=5pt]},
        line width=1.2pt,
        draw=black
    }
]

    \node[agentNode] (aWriter) {
        \textbf{Writer}\\[-1.5pt] {\footnotesize $a_{\text{writer}}$}
    };

    \node[paramNode, above left=0.75cm and 0.35cm of aWriter] (pTopic) {
        \textbf{topic}\\[-1.5pt] {\footnotesize \texttt{"Rocket"}}
    };
    \node[paramNode, above right=0.75cm and 0.35cm of aWriter] (pAge) {
        \textbf{target\_age}\\[-1.5pt] {\footnotesize 3\text{--}12}
    };

    \node[fileNode, below=0.85cm of aWriter] (fArticle) {
        \textbf{01\_draft.md}\\[-1.5pt] {\footnotesize Prose + placeholder}
    };

    \node[agentNode, below left=0.45cm and 0.9cm of fArticle] (aFact) {
        \textbf{Fact Checker}\\[-1.5pt] {\footnotesize $a_{\text{fact}}$}
    };
    \node[agentNode, below right=0.45cm and 0.9cm of fArticle] (aAge) {
        \textbf{Age Checker}\\[-1.5pt] {\footnotesize $a_{\text{age}}$}
    };

    \node[fileNode, below=0.85cm of aFact] (fAudit) {
        \textbf{fact\_audit.md}\\[-1.5pt] {\footnotesize Factual critique}
    };
    \node[fileNode, below=0.85cm of aAge] (fAgeRev) {
        \textbf{age\_audit.md}\\[-1.5pt] {\footnotesize Readability critique}
    };

    \node[gateNode, below=2.55cm of fArticle] (gate) {
        \textbf{Human Review Gate}\\[-1.5pt] {\footnotesize Editorial approval}
    };

    \node[agentNode, below=0.85cm of gate] (aIllust) {
        \textbf{Illustrator}\\[-1.5pt] {\footnotesize $a_{\text{illustrator}}$}
    };

    \node[fileNode, below=0.85cm of aIllust] (fPng) {
        \textbf{02\_illustration.png}\\[-1.5pt] {\footnotesize Synthesized image}
    };

    \node[font=\small\bfseries, below=0.85cm of fPng, xshift=-3.75cm] (legLbl) {Legend:};
    \node[paramNode, font=\footnotesize, inner xsep=3.5pt, inner ysep=2.2pt, right=0.25cm of legLbl] (legP) {Parameter $\mathcal{P}$};
    \node[agentNode, font=\footnotesize, minimum width=0pt, minimum height=0pt, rounded corners=4pt, inner xsep=5pt, inner ysep=2.5pt, right=0.30cm of legP] (legA) {Agent $\mathcal{A}$};
    \node[fileNode, font=\footnotesize, minimum width=0pt, minimum height=0pt, inner xsep=5pt, inner ysep=2.5pt, right=0.30cm of legA] (legF) {File $\mathcal{F}$};
    \node[gateNode, font=\footnotesize, minimum width=0pt, minimum height=0pt, inner xsep=5pt, inner ysep=3.2pt, right=0.30cm of legF] (legG) {\textit{Gate}};
    \begin{scope}[on background layer]
        \node[draw=black!50, fill=white, rounded corners=4pt, fit=(legLbl)(legP)(legA)(legF)(legG), inner sep=4.5pt] (legend) {};
    \end{scope}

    \draw[arrow] (pTopic) -- (aWriter);
    \draw[arrow] (pAge)   -- (aWriter);
    \draw[arrow, overlay] (pTopic) to[bend right=44] (aFact);
    \draw[arrow, overlay] (pAge) to[bend left=44] (aAge);

    \draw[arrow] (aWriter) -- (fArticle);
    \draw[arrow] (fArticle) -- (aFact);
    \draw[arrow] (fArticle) -- (aAge);

    \draw[cycleArrow] (aFact) -- (fAudit);
    \draw[cycleArrow] (aAge) -- (fAgeRev);

    \draw[cycleArrow, overlay] (fAudit) to[out=180, in=180, looseness=1.15] (aWriter);
    \draw[cycleArrow, overlay] (fAgeRev) to[out=0, in=0, looseness=1.15] (aWriter);

    \draw[arrow, line width=1.3pt] (fArticle) -- (gate);
    \draw[arrow] (gate) -- (aIllust);
    \draw[arrow] (aIllust) -- (fPng);

\end{tikzpicture}
\caption{Induced static tripartite workflow graph $G = (V, E)$ for \textsc{KidsWiki}, with distinct node shapes for parameters $\langle \mathcal{P} \rangle$, worker agents $(\mathcal{A})$, and file artifacts $[\mathcal{F}]$. The editorial review loop forms directed cycles.}
\label{fig:kidswiki_static}
\end{figure}
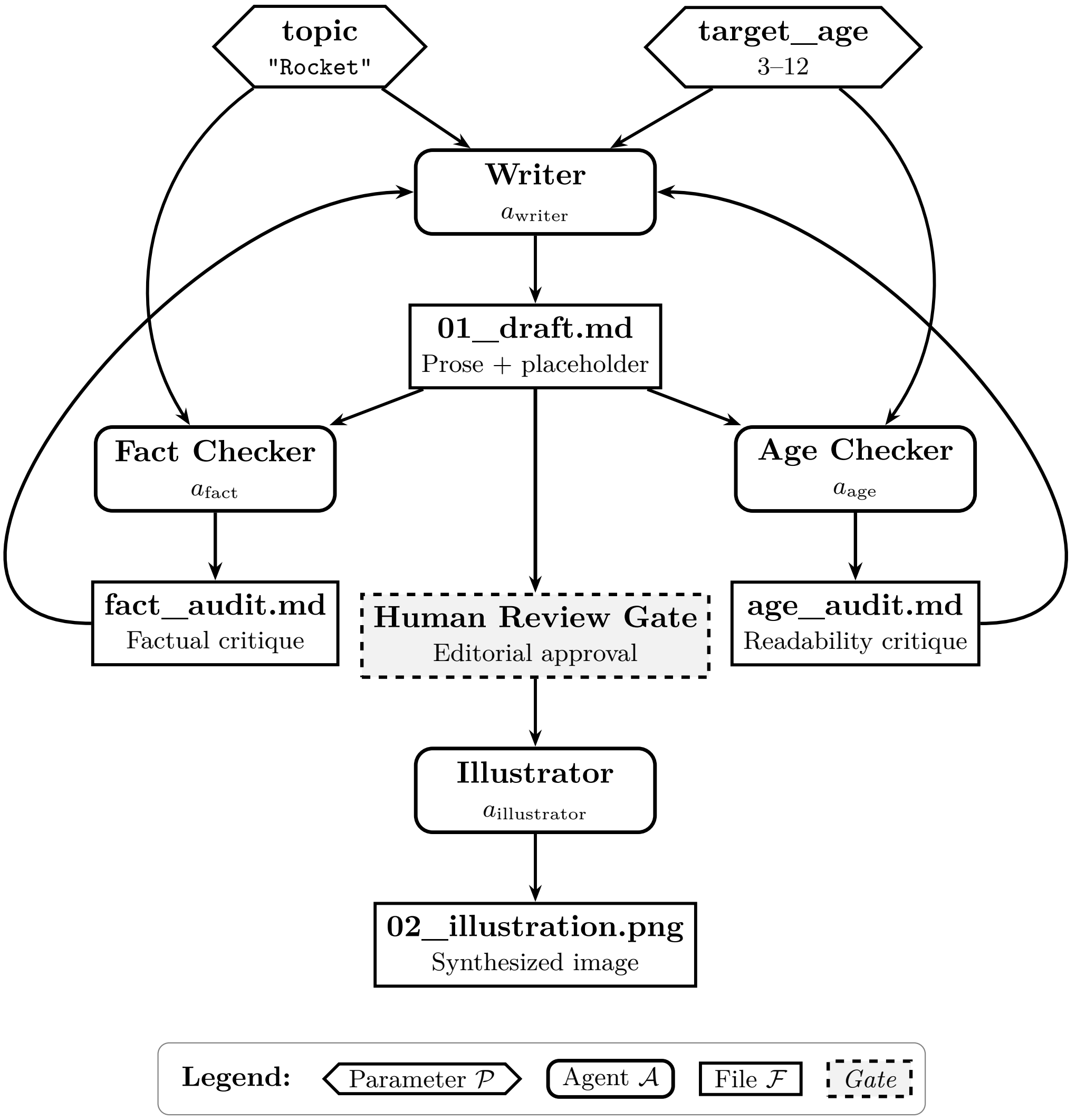

Because only agents compute while files and parameters hold passive state, every directed edge in $E \subseteq V \times V$ connects either an artifact to an agent (a permissible read) or an agent to an artifact (a permissible write):
\begin{equation}
    E = \bigcup_{a \in \mathcal{A}} \Big( \underbrace{\{(x, a) \mid x \in F_{\text{in}}(a) \cup P_{\text{in}}(a)\}}_{\text{Permissible Read Edges}} \;\cup\; \underbrace{\{(a, y) \mid y \in F_{\text{out}}(a) \cup P_{\text{out}}(a)\}}_{\text{Permissible Write Edges}} \Big)
\end{equation}
Iterative workflows such as generator--verifier loops require agents to exchange drafts and critiques across rounds, so $G$ is generally a \emph{directed cyclic graph}. In Figure~\ref{fig:kidswiki_static}, $a_{\text{writer}}$ writes \texttt{01\_draft.md} for $a_{\text{fact}}$ and $a_{\text{age}}$, which in turn write \texttt{fact\_audit.md} and \texttt{age\_audit.md} back to $a_{\text{writer}}$, forming two directed cycles:
\begin{align}
    a_{\text{writer}} \;\longrightarrow\; \texttt{01\_draft.md} \;\longrightarrow\; a_{\text{fact}} \;\longrightarrow\; \texttt{fact\_audit.md} \;\longrightarrow\; a_{\text{writer}} \\
    a_{\text{writer}} \;\longrightarrow\; \texttt{01\_draft.md} \;\longrightarrow\; a_{\text{age}} \;\longrightarrow\; \texttt{age\_audit.md} \;\longrightarrow\; a_{\text{writer}}
\end{align}

\subsection{Parallel Execution and the Unwound DAG \texorpdfstring{$\hat{G} = (\hat{V}, \hat{E})$}{G-hat = (V-hat, E-hat)}}
\label{sec:formal_unwound_dag}

\begin{figure}[p]
\centering
\begin{tikzpicture}[
    font=\normalsize,
    paramNode/.style={
        signal,
        signal to=east and west,
        draw=black,
        line width=1.05pt,
        fill=white,
        inner sep=3.8pt,
        align=center
    },
    agentNode/.style={
        rectangle,
        rounded corners=5pt,
        draw=black,
        line width=1.15pt,
        fill=white,
        minimum width=2.95cm,
        minimum height=0.84cm,
        align=center,
        inner xsep=2.5pt,
        inner ysep=3pt
    },
    fileNode/.style={
        rectangle,
        draw=black,
        line width=1.05pt,
        fill=white,
        minimum width=2.95cm,
        minimum height=0.80cm,
        align=center,
        inner xsep=2.5pt,
        inner ysep=3pt
    },
    gateNode/.style={
        rectangle,
        draw=black,
        dashed,
        fill=black!5,
        line width=1.15pt,
        minimum width=3.9cm,
        minimum height=0.84cm,
        align=center,
        inner sep=3pt
    },
    arrow/.style={
        -{Stealth[length=5.5pt, width=4pt]},
        line width=1.0pt,
        draw=black
    },
    cycleArrow/.style={
        -{Stealth[length=5.5pt, width=4pt]},
        line width=1.05pt,
        draw=black
    },
    gateArrow/.style={
        -{Stealth[length=6pt, width=4.5pt]},
        line width=1.2pt,
        draw=black
    },
    timeAxis/.style={
        -{Stealth[length=6pt, width=4.5pt]},
        line width=1.1pt,
        draw=black
    }
]

    \node[agentNode] (aW1) {
        {\small\textbf{Writer} (iter 1)}\\[-2pt]
        {\scriptsize $\alpha_1 = a_{\text{writer}}^{[0, t_1]}$}
    };

    \node[paramNode, above left=0.65cm and 0.35cm of aW1] (pTopic) {
        {\small\textbf{topic}$^{(0)}$}\\[-2pt]
        {\scriptsize \texttt{"Rocket"}}
    };
    \node[paramNode, above right=0.65cm and 0.35cm of aW1] (pAge) {
        {\small\textbf{target\_age}$^{(0)}$}\\[-2pt]
        {\scriptsize $4$}
    };

    \node[fileNode, below=0.68cm of aW1] (fA1) {
        {\small\textbf{01\_draft.md}$^{(t_1)}$}\\[-2pt]
        {\scriptsize Prose (\textit{``interstellar''} claim)}
    };

    \node[agentNode, below left=0.55cm and 0.20cm of fA1] (aF1) {
        {\small\textbf{Fact Checker} (iter 1)}\\[-2pt]
        {\scriptsize $\alpha_2 = a_{\text{fact}}^{[t_1, t_2]}$}
    };
    \node[agentNode, below right=0.55cm and 0.20cm of fA1] (aA1) {
        {\small\textbf{Age Checker} (iter 1)}\\[-2pt]
        {\scriptsize $\alpha_3 = a_{\text{age}}^{[t_1, t_2]}$}
    };

    \node[fileNode, below=0.68cm of aF1] (u1) {
        {\small\textbf{fact\_audit.md}$^{(t_2)}$}\\[-2pt]
        {\scriptsize \textbf{FAIL:} interstellar claim}
    };
    \node[fileNode, below=0.68cm of aA1] (r1) {
        {\small\textbf{age\_audit.md}$^{(t_2)}$}\\[-2pt]
        {\scriptsize \textbf{FAIL:} vocabulary}
    };

    \node[agentNode, below=3.55cm of fA1] (aW2) {
        {\small\textbf{Writer} (iter 2)}\\[-2pt]
        {\scriptsize $\alpha_4 = a_{\text{writer}}^{[t_2, t_3]}$}
    };

    \node[fileNode, below=0.68cm of aW2] (fA2) {
        {\small\textbf{01\_draft.md}$^{(t_3)}$}\\[-2pt]
        {\scriptsize Prose (\textit{``orbit \& Moon, simple''})}
    };

    \node[agentNode, below left=0.55cm and 0.20cm of fA2] (aF2) {
        {\small\textbf{Fact Checker} (iter 2)}\\[-2pt]
        {\scriptsize $\alpha_5 = a_{\text{fact}}^{[t_3, t_4]}$}
    };
    \node[agentNode, below right=0.55cm and 0.20cm of fA2] (aA2) {
        {\small\textbf{Age Checker} (iter 2)}\\[-2pt]
        {\scriptsize $\alpha_6 = a_{\text{age}}^{[t_3, t_4]}$}
    };

    \node[fileNode, below=0.68cm of aF2] (u2) {
        {\small\textbf{fact\_audit.md}$^{(t_4)}$}\\[-2pt]
        {\scriptsize \textbf{PASS:} verified facts}
    };
    \node[fileNode, below=0.68cm of aA2] (r2) {
        {\small\textbf{age\_audit.md}$^{(t_4)}$}\\[-2pt]
        {\scriptsize \textbf{PASS:} age 4 verified}
    };

    \node[gateNode, below=3.55cm of fA2] (gate) {
        {\small\textbf{Human Review Gate}}\\[-2pt]
        {\scriptsize Editorial approval (unlocked on PASS)}
    };

    \node[agentNode, below=0.72cm of gate] (aIL) {
        {\small\textbf{Illustrator}}\\[-2pt]
        {\scriptsize $\alpha_7 = a_{\text{illustrator}}^{[t_5, t_6]}$}
    };

    \node[fileNode, below=0.68cm of aIL] (fPng) {
        {\small\textbf{02\_illustration.png}$^{(t_6)}$}\\[-2pt]
        {\scriptsize Synthesized image}
    };

    \node[font=\small\bfseries, below=0.85cm of fPng, xshift=-3.75cm] (legLbl) {Legend:};
    \node[paramNode, font=\footnotesize, inner xsep=3.5pt, inner ysep=2.2pt, right=0.25cm of legLbl] (legP) {Parameter $\hat{\mathcal{P}}$};
    \node[agentNode, font=\footnotesize, minimum width=0pt, minimum height=0pt, rounded corners=4pt, inner xsep=5pt, inner ysep=2.5pt, right=0.30cm of legP] (legA) {Agent $\hat{\mathcal{A}}$};
    \node[fileNode, font=\footnotesize, minimum width=0pt, minimum height=0pt, inner xsep=5pt, inner ysep=2.5pt, right=0.30cm of legA] (legF) {File $\hat{\mathcal{F}}$};
    \node[gateNode, font=\footnotesize, minimum width=0pt, minimum height=0pt, inner xsep=5pt, inner ysep=3.2pt, right=0.30cm of legF] (legG) {\textit{Gate}};
    \begin{scope}[on background layer]
        \node[draw=black!50, fill=white, rounded corners=4pt, fit=(legLbl)(legP)(legA)(legF)(legG), inner sep=4.5pt] (legend) {};
    \end{scope}

    \draw[arrow] (pTopic) -- (aW1);
    \draw[arrow] (pTopic) to[bend right=20] (aF1);
    \draw[arrow] (pTopic) to[out=180, in=180, looseness=0.34] (aF2);
    \draw[arrow] (pAge) -- (aW1);
    \draw[arrow] (pAge) to[bend left=20] (aA1);
    \draw[arrow] (pAge) to[out=0, in=0, looseness=0.34] (aA2);

    \draw[arrow] (aW1) -- (fA1);
    \draw[arrow] (fA1) -- (aF1);
    \draw[arrow] (fA1) -- (aA1);

    \draw[arrow] (aF1) -- (u1);
    \draw[arrow] (aA1) -- (r1);

    \draw[cycleArrow] (u1) -- (aW2);
    \draw[cycleArrow] (r1) -- (aW2);

    \draw[arrow] (aW2) -- (fA2);
    \draw[arrow] (fA2) -- (aF2);
    \draw[arrow] (fA2) -- (aA2);

    \draw[arrow] (aF2) -- (u2);
    \draw[arrow] (aA2) -- (r2);

    \draw[gateArrow] (fA2) -- (gate);

    \draw[arrow] (gate) -- (aIL);
    \draw[arrow] (aIL) -- (fPng);

    \coordinate (axisTop) at ($(aF1.west |- pTopic)+(-1.10cm, 0.55cm)$);
    \coordinate (axisBot) at ($(aF1.west |- fPng)+(-1.10cm, -0.50cm)$);

    \draw[timeAxis] (axisTop)
        node[above=2pt, font=\footnotesize\bfseries, align=center] {Time $t \in \mathbb{R}_{\ge 0}$}
        -- (axisBot);

    \coordinate (rightBound) at ($(aA1.east)+(1.15cm, 0)$);
    \foreach \refNode/\tLabel in {
        pTopic/{$t_0=0$},
        fA1/{$t_1$},
        u1/{$t_2$},
        fA2/{$t_3$},
        u2/{$t_4$},
        gate/{$t_5$},
        fPng/{$t_6$}
    } {
        \coordinate (tk) at (axisTop |- \refNode);
        \begin{scope}[on background layer]
            \draw[densely dotted, black!30] (tk) -- (rightBound |- \refNode);
        \end{scope}
        \draw[black, line width=1.0pt] ($(tk)+(-3.5pt,0)$) -- ($(tk)+(3.5pt,0)$);
        \node[left=3.5pt, font=\footnotesize\bfseries] at (tk) {\tLabel};
    }

\end{tikzpicture}
\caption{Dynamic unwinding of the \textsc{KidsWiki} workflow into the execution DAG $\hat{G} = (\hat{V}, \hat{E})$ over runtime $t$. Invocations $\alpha_1 \dots \alpha_7$ create immutable snapshots $\texttt{01\_draft.md}^{(t)}$, $\texttt{fact\_audit.md}^{(t)}$, and $\texttt{age\_audit.md}^{(t)}$, unfolding static cycles into a strict topological order $\tau(u) < \tau(v)$. Once $\alpha_5$ and $\alpha_6$ pass at $t_4$, the orchestrator advances $\texttt{01\_draft.md}^{(t_3)}$ to editorial approval at $t_5$ ahead of Stage~2 illustration ($\alpha_7$).}
\label{fig:kidswiki_dag}
\end{figure}
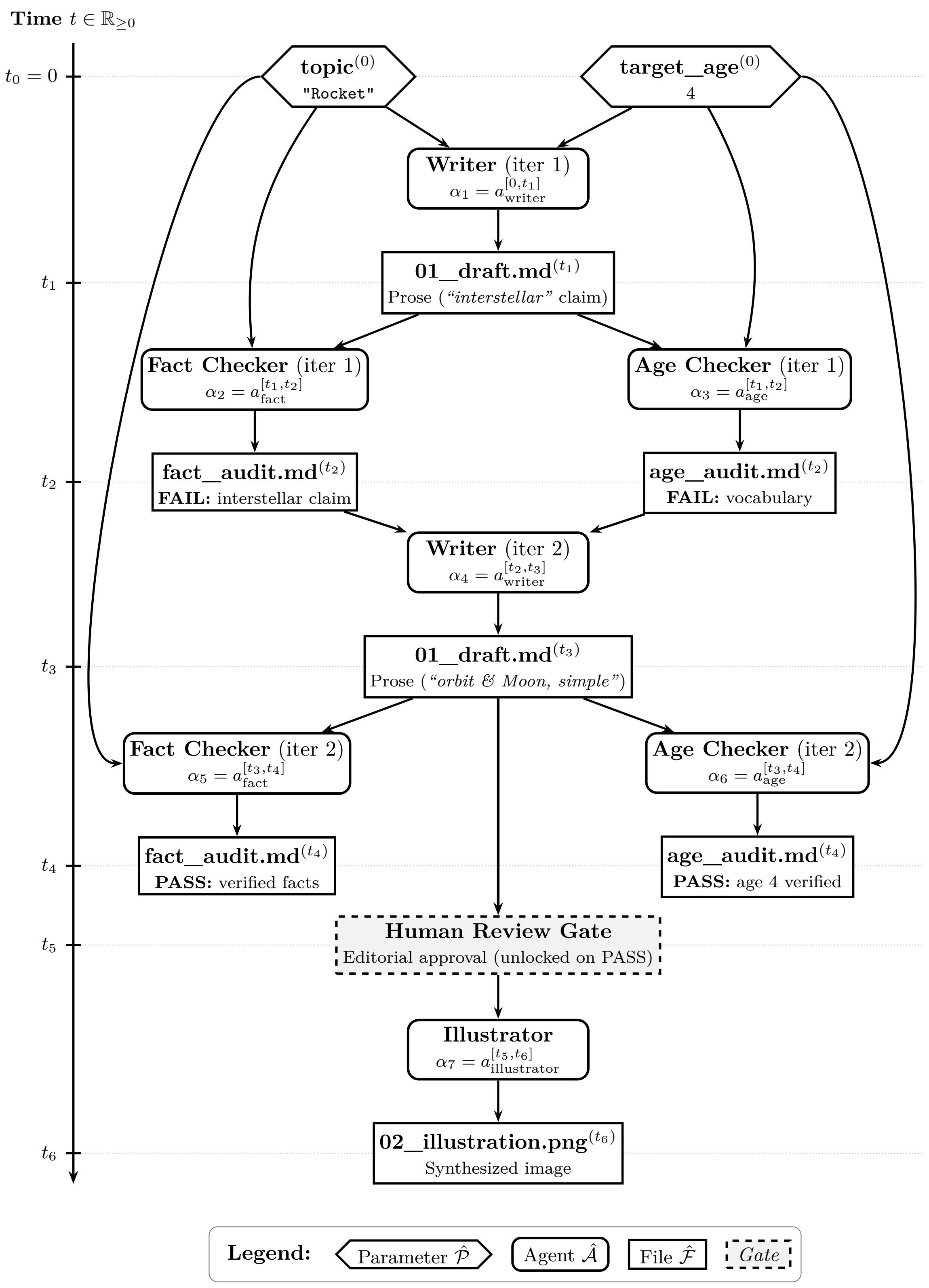

As execution progresses over runtime $t \in \mathbb{R}_{\ge 0}$, a scheduling \emph{Orchestrator} $\mathcal{O}$ unwinds the cyclic specification $G$ into a growing Directed Acyclic Graph $\hat{G} = (\hat{V}, \hat{E})$ representing the concrete provenance trace (Figure~\ref{fig:kidswiki_dag}).

\paragraph{Time-Indexed Vertices ($\hat{V}$).}
The unwound vertex set $\hat{V} = \hat{\mathcal{A}} \cup \hat{\mathcal{F}} \cup \hat{\mathcal{P}}$ records immutable artifact snapshots and bounded-duration agent invocations:
\begin{itemize}
    \item \textbf{State Snapshots ($\hat{\mathcal{F}}, \hat{\mathcal{P}}$):} $f^{(t)} \in \hat{\mathcal{F}}$ and $p^{(t)} \in \hat{\mathcal{P}}$ represent immutable versions of file $f \in \mathcal{F}$ and parameter $p \in \mathcal{P}$ created at time $t$. Execution starts at $t = 0$ from seed vertices $\hat{\mathcal{F}}_0 \cup \hat{\mathcal{P}}_0$ (for example, $\texttt{topic}^{(0)} = \text{``Rocket''}$ and $\texttt{target\_age}^{(0)} = 4$).
    \item \textbf{Parallel Agent Invocations ($\hat{\mathcal{A}}$):} $\alpha = a^{[t, t+n]} \in \hat{\mathcal{A}}$ denotes an invocation of agent $a \in \mathcal{A}$ started by $\mathcal{O}$ at time $t$ that runs for duration $n > 0$ and finishes at $t + n$. Two invocations $\alpha_1 = a_1^{[t_1, t_1+n_1]}$ and $\alpha_2 = a_2^{[t_2, t_2+n_2]}$ execute \emph{in parallel} whenever $[t_1, t_1+n_1] \cap [t_2, t_2+n_2] \neq \emptyset$.
\end{itemize}

\paragraph{Selective Subset I/O.}
An agent's static signature fixes the maximal set of ports it may access, whereas any single invocation $\alpha = a^{[t, t+n]} \in \hat{\mathcal{A}}$ reads and writes an \emph{active I/O footprint} contained within that boundary:
\begin{align}
    F_{\text{in}}(\alpha) &\subseteq F_{\text{in}}(a), & P_{\text{in}}(\alpha) &\subseteq P_{\text{in}}(a) \\
    F_{\text{out}}(\alpha) &\subseteq F_{\text{out}}(a), & P_{\text{out}}(\alpha) &\subseteq P_{\text{out}}(a)
\end{align}
On the first pass of \textsc{KidsWiki}, $\alpha_1 = a_{\text{writer}}^{[0, t_1]}$ reads only $\texttt{topic}^{(0)}$ and $\texttt{target\_age}^{(0)}$ ($F_{\text{in}}(\alpha_1) = \emptyset$, since no critique exists yet) and writes $\texttt{01\_draft.md}^{(t_1)}$. After parallel verification produces $\texttt{fact\_audit.md}^{(t_2)}$ and $\texttt{age\_audit.md}^{(t_2)}$, $\alpha_4 = a_{\text{writer}}^{[t_2, t_3]}$ incorporates both critiques to produce $\texttt{01\_draft.md}^{(t_3)}$. Once verifiers $\alpha_5$ and $\alpha_6$ return clean reviews at $t_4$, the orchestrator $\mathcal{O}$ advances $\texttt{01\_draft.md}^{(t_3)}$ directly to the human review gate at $t_5$ rather than relying on the writer to self-approve. Upon approval, $\alpha_7 = a_{\text{illustrator}}^{[t_5, t_6]}$ reads $\texttt{01\_draft.md}^{(t_3)}$ and generates $\texttt{02\_illustration.png}^{(t_6)}$.

\paragraph{Causal Edge Instantiation ($\hat{E}$).}
Only active ports of an invocation $\alpha = a^{[t, t+n]} \in \hat{\mathcal{A}}$ add vertices and edges to $\hat{G}$:
\begin{enumerate}
    \item \textbf{Selective Input Binding (at start time $t$):} Each active input $x \in F_{\text{in}}(\alpha) \cup P_{\text{in}}(\alpha)$ is bound to an existing snapshot $x^{(t')} \in \hat{\mathcal{F}} \cup \hat{\mathcal{P}}$ with $t' \le t$ (by default the latest snapshot $t' = \max\{\tau \le t \mid x^{(\tau)} \in \hat{V}\}$):
    \begin{equation}
        (x^{(t')}, \; a^{[t, t+n]}) \in \hat{E} \quad \forall x \in F_{\text{in}}(\alpha) \cup P_{\text{in}}(\alpha)
    \end{equation}
    \item \textbf{Selective Output Materialization (at completion time $t+n$):} At $t+n$, new immutable snapshot vertices $y^{(t+n)} \in \hat{\mathcal{F}} \cup \hat{\mathcal{P}}$ are instantiated if and only if $y \in F_{\text{out}}(\alpha) \cup P_{\text{out}}(\alpha)$, connected by write edges:
    \begin{equation}
        (a^{[t, t+n]}, \; y^{(t+n)}) \in \hat{E} \quad \forall y \in F_{\text{out}}(\alpha) \cup P_{\text{out}}(\alpha)
    \end{equation}
\end{enumerate}

\subsection{Structural Execution Invariants}

Connecting $G$ and $\hat{G}$ gives two straightforward invariants: runtime traces remain strictly acyclic (Proposition~1) and never step outside the static capability surface declared in $G$ (Proposition~2).

\medskip
\noindent\textbf{Proposition 1 (Strict Acyclicity of $\hat{G}$).} \emph{For any parallel execution trace with positive invocation durations ($n > 0$), the unwound graph $\hat{G} = (\hat{V}, \hat{E})$ is a Directed Acyclic Graph (DAG).}

\emph{Proof.} Define $\tau: \hat{V} \to \mathbb{R}_{\ge 0}$ by $\tau(x^{(t)}) = t$ for state snapshots $x^{(t)} \in \hat{\mathcal{F}} \cup \hat{\mathcal{P}}$ and $\tau(a^{[t, t+n]}) = t + \frac{n}{2}$ for invocations $a^{[t, t+n]} \in \hat{\mathcal{A}}$. Along every input edge $(x^{(t')}, a^{[t, t+n]}) \in \hat{E}$, $t' \le t$ and $n > 0$ imply $\tau(x^{(t')}) = t' \le t < t + \frac{n}{2} = \tau(a^{[t, t+n]})$. Along every output edge $(a^{[t, t+n]}, y^{(t+n)}) \in \hat{E}$, $\tau(a^{[t, t+n]}) = t + \frac{n}{2} < t + n = \tau(y^{(t+n)})$. Since $\tau(u) < \tau(v)$ holds strictly on every edge $(u, v) \in \hat{E}$, $\hat{G}$ contains no directed cycles. \hfill $\square$

\medskip
\noindent\textbf{Proposition 2 (Subgraph-Compliant Projection Homomorphism).} \emph{Let $\pi: \hat{V} \to V$ be the type-preserving projection $\pi(a^{[t, t+n]}) = a$, $\pi(f^{(t)}) = f$, and $\pi(p^{(t)}) = p$. Then $\pi$ is a graph homomorphism from $\hat{G}$ to $G$.}

\emph{Proof.} Every edge $(u, v) \in \hat{E}$ is either an input edge $(x^{(t')}, a^{[t, t+n]})$ with $x \in F_{\text{in}}(\alpha) \cup P_{\text{in}}(\alpha) \subseteq F_{\text{in}}(a) \cup P_{\text{in}}(a)$ or an output edge $(a^{[t, t+n]}, y^{(t+n)})$ with $y \in F_{\text{out}}(\alpha) \cup P_{\text{out}}(\alpha) \subseteq F_{\text{out}}(a) \cup P_{\text{out}}(a)$. In both cases $(\pi(u), \pi(v)) \in E$, so every runtime trace $\hat{G}$ respects the static read/write boundaries of $G$. \hfill $\square$

\paragraph{Remark on the Orchestrator ($\mathcal{O}$) and Human Review Gates.}
In the formal model, both the orchestrator $\mathcal{O}$ and human review gates (shown as dashed boxes in Figures~\ref{fig:kidswiki_static} and~\ref{fig:kidswiki_dag}) reside in the control plane rather than in $V = \mathcal{A} \cup \mathcal{F} \cup \mathcal{P}$. $\mathcal{O}$ inspects the frontier of $\hat{G}$, pauses at review gates until an operator approves an existing snapshot $f^{(t)}$ (such as $\texttt{01\_draft.md}^{(t_3)}$ at $t_5$), and dispatches downstream invocations without mutating artifacts directly. If an operator requests changes at a gate, $\mathcal{O}$ routes the revision back to the responsible worker session so every file write remains an explicit provenance edge in $\hat{G}$.

\section{\weave{} Implementation}
\label{sec:implementation}

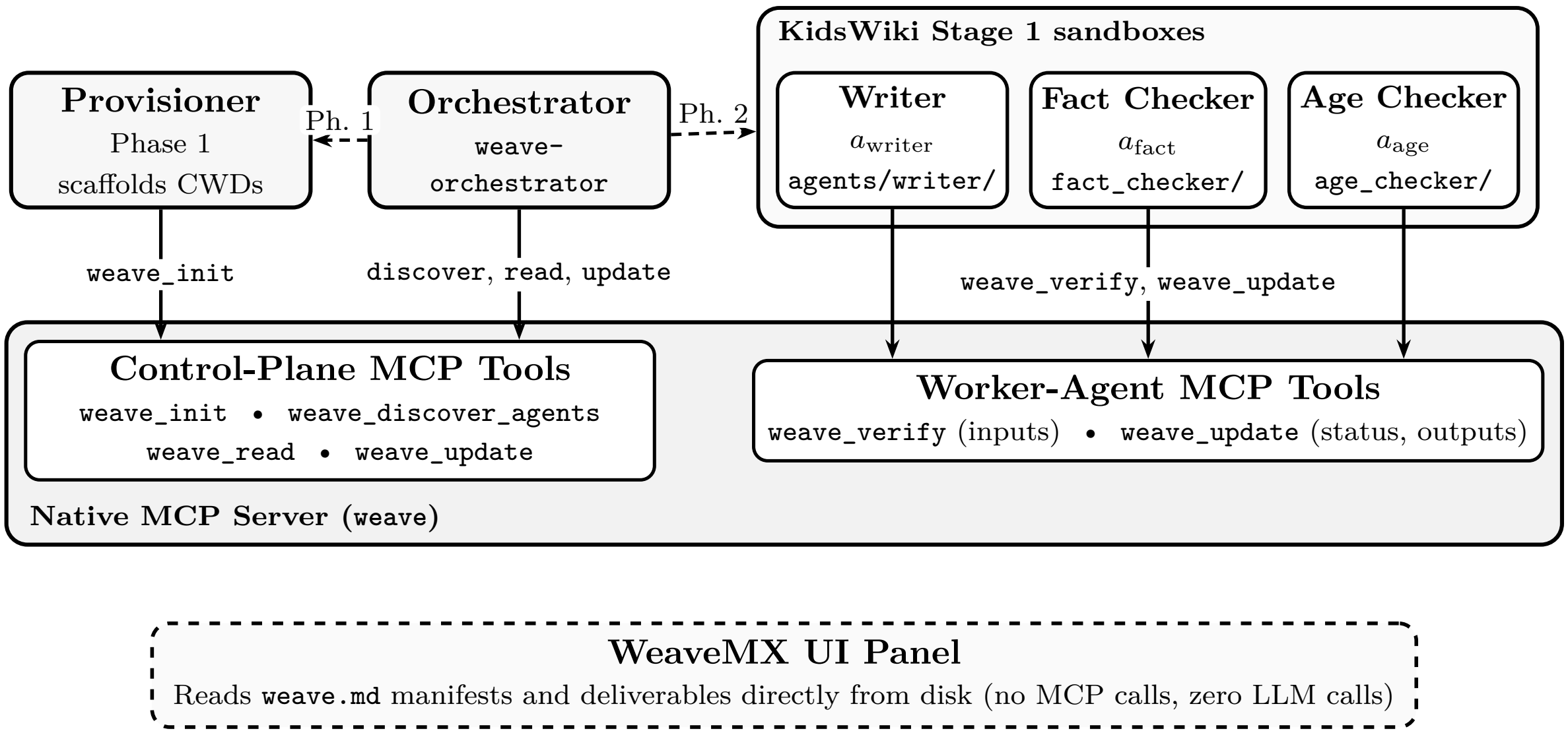
\begin{figure*}[t]
\centering
\begin{tikzpicture}[
    font=\normalsize,
    controlNode/.style={
        rectangle, rounded corners=5pt, draw=black, fill=black!3,
        line width=1.15pt, align=center, minimum width=3.0cm,
        minimum height=1.35cm, inner xsep=4pt, inner ysep=4pt
    },
    agentSandbox/.style={
        rectangle, rounded corners=5pt, draw=black, fill=white,
        line width=1.0pt, align=center, minimum width=2.15cm,
        minimum height=1.35cm, inner xsep=3pt, inner ysep=3.5pt
    },
    groupBox/.style={
        rectangle, rounded corners=6pt, draw=black, fill=black!2, line width=1.1pt
    },
    mcpGroup/.style={
        rectangle, rounded corners=6pt, draw=black, fill=black!5, line width=1.15pt
    },
    mcpSub/.style={
        rectangle, rounded corners=4pt, draw=black, fill=white,
        line width=0.95pt, align=center, inner xsep=4pt, inner ysep=4pt
    },
    sidecarNode/.style={
        rectangle, rounded corners=5pt, draw=black, dashed, fill=black!2,
        line width=1.05pt, align=center, inner xsep=6pt, inner ysep=4.5pt
    },
    arrow/.style={-{Stealth[length=5.5pt, width=4pt]}, line width=1.0pt, draw=black},
    dispatchArrow/.style={-{Stealth[length=5.5pt, width=4pt]}, line width=0.95pt, densely dashed, draw=black},
    edgeLabel/.style={font=\scriptsize, fill=white, inner sep=1.5pt, rounded corners=2pt}
]

    \node[controlNode] (prov) {
        {\small\textbf{Provisioner}}\\[-2pt]
        {\scriptsize Phase 1}\\[-2pt]
        {\scriptsize scaffolds CWDs}
    };
    \node[controlNode, right=0.55cm of prov] (orch) {
        {\small\textbf{Orchestrator}}\\[-2pt]
        {\scriptsize \texttt{weave-}}\\[-3pt]
        {\scriptsize \texttt{orchestrator}}
    };

    \node[agentSandbox, right=1.05cm of orch] (wWriter) {
        {\footnotesize\textbf{Writer}}\\[-2pt]
        {\scriptsize $a_{\text{writer}}$}\\[-2pt]
        {\scriptsize \texttt{agents/writer/}}
    };
    \node[agentSandbox, right=0.2cm of wWriter] (wFact) {
        {\footnotesize\textbf{Fact Checker}}\\[-2pt]
        {\scriptsize $a_{\text{fact}}$}\\[-2pt]
        {\scriptsize \texttt{fact\_checker/}}
    };
    \node[agentSandbox, right=0.2cm of wFact] (wAge) {
        {\footnotesize\textbf{Age Checker}}\\[-2pt]
        {\scriptsize $a_{\text{age}}$}\\[-2pt]
        {\scriptsize \texttt{age\_checker/}}
    };
    \coordinate (wTopPad) at ($(wFact.north)+(0, 13pt)$);
    \begin{scope}[on background layer]
        \node[groupBox, fit=(wWriter)(wFact)(wAge)(wTopPad), inner sep=5pt] (s1Box) {};
    \end{scope}
    \node[anchor=north west, font=\scriptsize\bfseries, inner sep=0pt] at ($(s1Box.north west)+(6pt,-4pt)$) {
        \textsc{KidsWiki} Stage 1 sandboxes
    };

    \coordinate (ctrlMid) at ($(prov)!0.5!(orch)$);
    \node[mcpSub, minimum width=6.8cm, minimum height=1.0cm, below=1.5cm of wFact] (mcpAgent) {
        {\small\textbf{Worker-Agent MCP Tools}}\\[-2pt]
        {\scriptsize \texttt{weave\_verify} (inputs) \;\textbullet\; \texttt{weave\_update} (status, outputs)}
    };
    \node[mcpSub, minimum width=6.3cm, minimum height=1.0cm] (mcpOrch) at (ctrlMid |- mcpAgent) {
        {\small\textbf{Control-Plane MCP Tools}}\\[-2pt]
        {\scriptsize \texttt{weave\_init} \;\textbullet\; \texttt{weave\_discover\_agents}}\\[-2.5pt]
        {\scriptsize \texttt{weave\_read} \;\textbullet\; \texttt{weave\_update}}
    };
    \coordinate (mcpBotPad) at ($(mcpOrch.south)+(0, -13pt)$);
    \begin{scope}[on background layer]
        \node[mcpGroup, fit=(mcpOrch)(mcpAgent)(mcpBotPad), inner sep=5pt] (mcpBox) {};
    \end{scope}
    \node[anchor=south west, font=\scriptsize\bfseries, inner sep=0pt] at ($(mcpBox.south west)+(7pt,4pt)$) {
        Native MCP Server (\texttt{weave})
    };

    \node[sidecarNode, below=0.75cm of mcpBox] (sidecar) {
        {\small\textbf{WeaveMX UI Panel}}\\[-2pt]
        {\scriptsize Reads \texttt{weave.md} manifests and deliverables directly from disk (no MCP calls, zero LLM calls)}
    };

    \draw[dispatchArrow] (orch) -- node[edgeLabel, above=1pt] {Ph.~1} (prov);
    \draw[dispatchArrow] (orch) -- node[edgeLabel, above=1pt] {Ph.~2} (s1Box);

    \draw[arrow] (prov) -- node[edgeLabel] {\texttt{weave\_init}} (prov |- mcpOrch.north);
    \draw[arrow] (orch) -- node[edgeLabel] {\texttt{discover}, \texttt{read}, \texttt{update}} (orch |- mcpOrch.north);

    \draw[arrow] (wWriter) -- (wWriter |- mcpAgent.north);
    \draw[arrow] (wFact) -- node[edgeLabel] {\texttt{weave\_verify}, \texttt{weave\_update}} (mcpAgent);
    \draw[arrow] (wAge) -- (wAge |- mcpAgent.north);

\end{tikzpicture}
\caption{\weave{} multi-layer architecture illustrated on Stage~1 of \textsc{KidsWiki}. Partially predefined Markdown stage and agent definitions are executed predictably via \texttt{weave-orchestrator} and \texttt{weave-agent} skills, backed by \mcpweave{} MCP tools (\texttt{weave\_init}, \texttt{weave\_discover\_agents}, \texttt{weave\_read}, and \texttt{weave\_update} on the control plane, and \texttt{weave\_verify} and \texttt{weave\_update} inside worker sandboxes) that enforce port contracts and manage lifecycles in on-disk \texttt{weave.md} manifests. An out-of-band UI polls those manifests with zero LLM calls.}
\label{fig:architecture}
\end{figure*}

To realize the formal model from Section~\ref{sec:formal_model} with zero orchestration code, \weave{} is implemented as an open-source, MIT-licensed multi-harness toolkit with zero external package dependencies (Figure~\ref{fig:architecture}), available at \url{https://github.com/axiomfab/weavemx}. Predictable execution follows a direct causal chain across three layers. First, orchestrators and worker agents are specified in partially predefined, declarative Markdown files. Second, \weave{} orchestrator and worker skills prescribe deterministic operating protocols so agents execute those Markdown specifications predictably. Third, the \mcpweave{} Model Context Protocol (MCP) server enforces those protocols programmatically on disk, validating port contracts and recording lifecycle transitions in \texttt{weave.md} manifests.

\subsection{Declarative Pipeline Authoring and Stage Orchestration}
\label{sec:impl_stages_authoring}

A \weave{} pipeline contains no Python or TypeScript orchestration code: both the control-plane stage roadmap and every specialist role are written in Markdown following a partially predefined declarative structure (combining YAML frontmatter and canonical section headers for roles, ports, parameters, tasks, and review gates with free-form domain instructions). Three skills connect authoring-time specification to predictable runtime execution and \texttt{weave.md} lifecycle tracking:
\begin{itemize}
    \item \textbf{Skill 1: \texttt{weave-authoring}} (\emph{Authoring Time}): Synthesizes and revises partially predefined Markdown agent definitions and stage guides from natural-language briefs or run feedback, backed by static data-flow auditing so specifications start inside $\Sigma_{\text{valid}}$.
    \item \textbf{Skill 2: \texttt{weave-orchestrator}} (\emph{Predictable Control-Plane Execution}): Ensures the root orchestrator and stage sub-orchestrators execute \texttt{orchestrator.md} and \texttt{stage*.md} through repeatable control-flow rules, chaining \texttt{orchestrator-agent-instantiation} (Phase~1 workspace provisioning) and \texttt{orchestrator-agent-execution} (Phase~2 worker dispatch and completion checks) while tracking pipeline and stage lifecycles in \texttt{weave.md} via the \mcpweave{} MCP server.
    \item \textbf{Skill 3: \texttt{weave-agent}} (\emph{Predictable Worker Execution}): Ensures worker subagents execute their \texttt{.agent/agents/*.md} specifications predictably inside isolated directories, requiring Turn-1 pre-flight input verification (\texttt{weave\_verify}), incremental subtask logging, and output-verified status transitions (\texttt{weave\_update}) recorded in each worker's \texttt{weave.md}.
\end{itemize}

\paragraph{Flat Agent Specifications (\texttt{.agent/agents/*.md}).}
Each worker agent $a \in \mathcal{A}$ is defined by a standalone Markdown instruction file in \texttt{.agent/agents/<agent\_name>.md}, which the host registers as an invocable subagent type that the orchestrator may instantiate one or more times. Its YAML frontmatter pins the model tier, attaches the \texttt{weave-agent} skill, and restricts built-in tools to those needed by the role (such as \texttt{view\_file} and \texttt{write\_to\_file}). In the Markdown body, \texttt{\# Role Specifics} states the agent's domain task and local \textbf{Output Target} filenames (e.g., \texttt{draft.md} or \texttt{segment\_*.md}), referring to inputs strictly by local alias so the agent stays decoupled from stage numbers and run paths. Single-purpose roles (such as an atomic verifier or image generator) need only \texttt{\# Role Specifics} and report \texttt{in\_progress} directly, whereas multi-step roles optionally add \texttt{\# Tasks} to break internal work into numbered \texttt{snake\_case} subtasks (\texttt{\#\# Task: 01\_outline}, \texttt{\#\# Task: 02\_draft}, mirrored in status prefixes such as \texttt{in\_progress/01\_outline}) and \texttt{\# Lifecycle} to specify subtask ordering and conditional branches without hardcoding low-level MCP tool calls or JSON schemas inside agent instructions.

\paragraph{Stage-Modular Guides and Two-Phase Execution.}
Multi-stage pipelines divide into sequential stages ($s_1, \dots, s_K$) described by a root guide (\texttt{orchestrator.md}) and per-stage files (\texttt{orchestrator/stage<N>\_<slug>.md}). Each stage file lists its leaf agents (\texttt{\#\# Agents}) with their instruction paths, working directories, explicit or aliased \texttt{Inputs}, and optional \texttt{Outputs}, followed by an \texttt{\#\# Execution} plan with explicit \texttt{Parameters:} lines and optional \textbf{User Review Gates}.

At runtime, the root orchestrator follows \texttt{weave-orchestrator}. In Step~0, it creates all stage folders under \texttt{runs/<context\_slug>/} and initializes the root \texttt{weave.md} manifest so the UI can render the full stage stepper right away. Each stage then runs in two phases (delegated per stage to an ephemeral \texttt{stage\_orchestrator} in multi-stage pipelines) to keep control-plane scheduling separate from workspace setup:
\begin{itemize}
    \item \textbf{Phase 1: Provisioner Delegation (\texttt{orchestrator-agent-instantiation}).} The root orchestrator never creates worker directories or worker \texttt{weave.md} files itself. At the start of Stage~$k$, it dispatches a lightweight \texttt{provisioner} subagent that reads \texttt{\#\# Agents}, creates each worker's isolated folder, calls \texttt{weave\_init} to populate its \texttt{weave.md} frontmatter (\texttt{instructions\_file}, \texttt{inputs}, \texttt{outputs}), and returns \texttt{PROVISIONING\_COMPLETE}.
    \item \textbf{Phase 2: Worker Dispatch and Execution Topologies (\texttt{orchestrator-agent-execution}).} Once provisioning finishes, the orchestrator queries \texttt{weave\_discover\_agents} (filtering for \texttt{initialized} nodes) and dispatches workers under one of two protocols:
    \begin{enumerate}
        \item \emph{Protocol A (Direct Sequential and Parallel Waves):} The orchestrator launches leaf subagents sequentially or in parallel batches via \texttt{invoke\_subagent}, passing only the working directory, explicit \texttt{Runtime Parameters}, and local \texttt{Required Output} filenames. In cyclic generator--verifier loops (such as Writer and Fact Checker), critique paths are sent back to the live generator session via \texttt{send\_message}, preserving its working memory across rounds.
        \item \emph{Protocol B (Hierarchical Sub-Orchestrators):} When a stage fans out across $M$ independent units that each need their own multi-agent sequence or revision loop (for example, drafting and checking ten book chapters in parallel), the root orchestrator spawns $M$ parallel \emph{sub-orchestrators}. Each sub-orchestrator runs the local loop and \texttt{weave\_read} checks for its unit inside the shared \texttt{runs/<context\_slug>/} tree without creating nested run roots or touching the root \texttt{weave.md}.
    \end{enumerate}
\end{itemize}

\paragraph{One-Shot Pipeline Synthesis and Constrained Revision (\texttt{weave-authoring}).}
Let $\Sigma_{\text{valid}}$ denote the space of \weave{} specifications $\sigma = \langle \mathcal{A}, \mathcal{S} \rangle$ (comprising flat agent definitions $\mathcal{A}$ and stage guides $\mathcal{S} = \langle s_1, \dots, s_K \rangle$) that satisfy the data-flow consistency checks in Section~\ref{sec:verification}. Because $\sigma$ is pure Markdown without code dependencies, \texttt{weave-authoring} can synthesize an initial valid specification $\sigma_0 \in \Sigma_{\text{valid}}$ from a short two-to-five-sentence brief in one pass.

Iterative pipeline development requires refining a specification after inspecting real outputs without breaking upstream or downstream contracts. When an operator submits feedback items $C = \{c_1, \dots, c_x\}$ on a completed run in \texttt{runs/<context\_slug>/}, \texttt{weave-authoring} performs structured feedback triage to map systemic issues to candidate specification edits $R = \{r_1, \dots, r_y\}$:
\begin{enumerate}
    \item \textbf{Triage ($C = C_{\text{taste}} \cup C_{\text{sys}}$):} Each feedback item $c_i$ is classified either as a \emph{Run-Specific Preference} ($c_i \in C_{\text{taste}}$, such as asking for a red rocket instead of a blue rocket on one \textsc{KidsWiki} run) or a \emph{Systemic Pipeline Defect} ($c_i \in C_{\text{sys}}$, such as noticing that \texttt{age\_checker} never audited the italicized image caption).
    \item \textbf{Root-Cause Mapping ($C_{\text{sys}} \to R$):} For each $c_i \in C_{\text{sys}}$, parallel read-only \texttt{research} subagents compare the run artifacts against $\sigma_k$ to propose minimal specification edits $R$ on concrete agent tasks or port bindings, while leaving one-off preferences $C_{\text{taste}}$ out of $R$.
    \item \textbf{Audited Update ($\sigma_{k+1} \in \Sigma_{\text{valid}}$):} After the user chooses whether to patch the current run, update the reusable specification $\sigma_k$, or do both, \texttt{weave-authoring} applies $R$ and re-runs the data-flow audit, rejecting any edit that would violate port matching or input availability.
\end{enumerate}

\subsection{Workspace Isolation and Scoped File Resolution}
\label{sec:impl_isolation}

To back the formal view of worker agents as bounded operators $a = \langle F_{\text{in}}, F_{\text{out}}, P_{\text{in}}, P_{\text{out}} \rangle$, each worker invocation runs in an isolated directory scope.

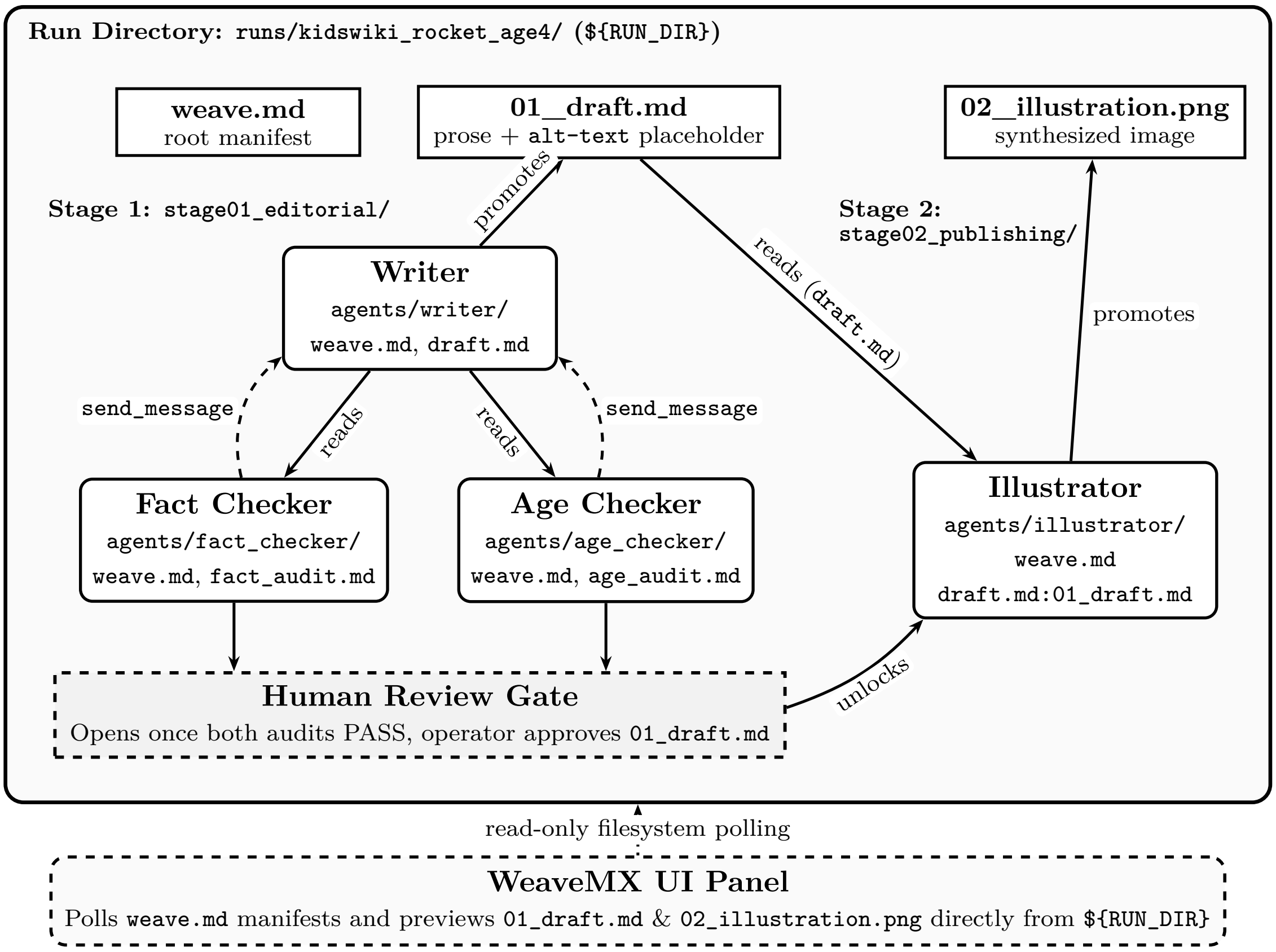
\begin{figure*}[t]
\centering
\begin{tikzpicture}[
    font=\normalsize,
    outerContainer/.style={
        rectangle,
        rounded corners=6pt,
        draw=black,
        line width=1.25pt,
        fill=black!2,
        align=center
    },
    stageContainer/.style={
        rectangle,
        rounded corners=5pt,
        draw=black,
        line width=1.05pt,
        fill=white,
        align=center
    },
    agentBox/.style={
        rectangle,
        rounded corners=4pt,
        draw=black,
        line width=1.0pt,
        fill=white,
        align=center,
        inner xsep=4pt,
        inner ysep=4.5pt
    },
    fileChip/.style={
        rectangle,
        draw=black,
        fill=white,
        line width=0.95pt,
        align=center,
        inner xsep=5pt,
        inner ysep=3.5pt,
        font=\footnotesize
    },
    gateBox/.style={
        rectangle,
        draw=black,
        dashed,
        line width=1.1pt,
        fill=black!5,
        align=center,
        inner xsep=5pt,
        inner ysep=4pt
    },
    sidecarBox/.style={
        rectangle,
        rounded corners=5pt,
        draw=black,
        dashed,
        line width=1.05pt,
        fill=black!2,
        align=center,
        inner sep=4.5pt
    },
    arrow/.style={
        -{Stealth[length=5pt, width=3.8pt]},
        line width=0.95pt,
        draw=black
    },
    cycleArrow/.style={
        -{Stealth[length=5pt, width=3.8pt]},
        line width=0.95pt,
        dashed,
        draw=black
    },
    syncArrow/.style={
        -{Stealth[length=5pt, width=3.8pt]},
        line width=0.95pt,
        dotted,
        draw=black
    },
    edgeLabel/.style={
        font=\scriptsize,
        fill=white,
        inner sep=1.4pt,
        rounded corners=2pt
    }
]

    \node[agentBox, minimum width=3.15cm] (fcBox) {
        {\small\textbf{Fact Checker}}\\[-2pt]
        {\scriptsize \texttt{agents/fact\_checker/}}\\[-2pt]
        {\scriptsize \texttt{weave.md}, \texttt{fact\_audit.md}}
    };
    \node[agentBox, minimum width=3.15cm, right=0.80cm of fcBox] (ageBox) {
        {\small\textbf{Age Checker}}\\[-2pt]
        {\scriptsize \texttt{agents/age\_checker/}}\\[-2pt]
        {\scriptsize \texttt{weave.md}, \texttt{age\_audit.md}}
    };
    \node[agentBox, minimum width=3.20cm, above=1.25cm of $(fcBox.north)!0.5!(ageBox.north)$] (wBox) {
        {\small\textbf{Writer}}\\[-2pt]
        {\scriptsize \texttt{agents/writer/}}\\[-2pt]
        {\scriptsize \texttt{weave.md}, \texttt{draft.md}}
    };

    \node[gateBox, minimum width=7.10cm, below=0.80cm of $(fcBox.south)!0.5!(ageBox.south)$] (gate) {
        {\small\textbf{Human Review Gate}}\\[-2pt]
        {\scriptsize Opens once both audits PASS, operator approves \texttt{01\_draft.md}}
    };

    \node[agentBox, minimum width=3.55cm, right=1.85cm of ageBox] (ilBox) {
        {\small\textbf{Illustrator}}\\[-2pt]
        {\scriptsize \texttt{agents/illustrator/}}\\[-2pt]
        {\scriptsize \texttt{weave.md}}\\[-2pt]
        {\scriptsize \texttt{draft.md:01\_draft.md}}
    };

    \node[fileChip, minimum width=2.85cm] (runWeave) at ($(fcBox |- wBox.north)+(0.05cm, 1.45cm)$) {
        \textbf{weave.md}\\[-2pt]
        {\scriptsize root manifest}
    };
    \node[fileChip, minimum width=4.05cm] (runArt) at ($(wBox.north)+(2.10cm, 1.45cm)$) {
        \textbf{01\_draft.md}\\[-2pt]
        {\scriptsize prose + \texttt{alt-text} placeholder}
    };
    \node[fileChip, minimum width=3.45cm] (runPng) at ($(ilBox |- runArt)+(0.35cm, 0)$) {
        \textbf{02\_illustration.png}\\[-2pt]
        {\scriptsize synthesized image}
    };

    \coordinate (s1Top) at ($(wBox.north)+(0, 14pt)$);
    \coordinate (s1LeftPad) at ($(fcBox.west)+(-10pt, 0)$);
    \coordinate (s1RightPad) at ($(ageBox.east)+(10pt, 0)$);
    \coordinate (s2Top) at ($(s1Top -| ilBox)$);
    \coordinate (s2Bot) at ($(gate.south -| ilBox)$);
    \coordinate (s2LeftPad) at ($(ilBox.west)+(-0.85cm, 0)$);

    \begin{scope}[on background layer]
        \node[stageContainer, fit=(wBox)(fcBox)(ageBox)(gate)(s1Top)(s1LeftPad)(s1RightPad), inner sep=6pt] (s1) {};
        \node[stageContainer, fit=(ilBox)(s2Top)(s2Bot)(s2LeftPad), inner sep=6pt] (s2) {};
        \coordinate (runTop) at ($(runArt.north)+(0, 18pt)$);
        \node[outerContainer, fit=(runWeave)(runArt)(runPng)(s1)(s2)(runTop), inner sep=8pt] (run) {};
    \end{scope}

    \node[anchor=north west, font=\scriptsize\bfseries, inner sep=0pt] at ($(run.north west)+(8pt,-5pt)$) {
        Run Directory: \texttt{runs/kidswiki\_rocket\_age4/} (\texttt{\$\{RUN\_DIR\}})
    };
    \node[anchor=north west, font=\scriptsize\bfseries, inner sep=0pt] at ($(s1.north west)+(6pt,-5pt)$) {
        Stage 1: \texttt{stage01\_editorial/}
    };
    \node[anchor=north west, align=left, font=\scriptsize\bfseries, inner sep=0pt] at ($(s2.north west)+(6pt,-5pt)$) {
        Stage 2:\\[-1.5pt]{\scriptsize\texttt{stage02\_publishing/}}
    };

    \node[sidecarBox, minimum width=12.4cm, below=0.6cm of run] (sidecar) {
        {\small\textbf{WeaveMX UI Panel}}\\[-2pt]
        {\scriptsize Polls \texttt{weave.md} manifests and previews \texttt{01\_draft.md} \& \texttt{02\_illustration.png} directly from \texttt{\$\{RUN\_DIR\}}}
    };

    \draw[arrow] (wBox) -- node[edgeLabel, above, sloped, pos=0.52] {promotes} (runArt);
    \draw[arrow] (wBox) -- node[edgeLabel, below, sloped, pos=0.48] {reads} (fcBox);
    \draw[arrow] (wBox) -- node[edgeLabel, below, sloped, pos=0.48] {reads} (ageBox);

    \draw[cycleArrow] (fcBox) to[bend left=32] node[edgeLabel, left=1pt, pos=0.5] {\texttt{send\_message}} (wBox);
    \draw[cycleArrow] (ageBox) to[bend right=32] node[edgeLabel, right=1pt, pos=0.5] {\texttt{send\_message}} (wBox);

    \draw[arrow] (fcBox) -- (fcBox |- gate.north);
    \draw[arrow] (ageBox) -- (ageBox |- gate.north);
    \draw[arrow] (gate) to[bend right=14] node[edgeLabel, below, sloped, pos=0.52] {unlocks} (ilBox);

    \draw[arrow] (runArt) -- node[edgeLabel, above, sloped, pos=0.52] {reads (\texttt{draft.md})} (ilBox);
    \draw[arrow] (ilBox) -- node[edgeLabel, right=2pt, pos=0.48] {promotes} (runPng);

    \draw[syncArrow] (sidecar) -- node[edgeLabel, fill=white] {read-only filesystem polling} (run);

\end{tikzpicture}
\caption{Physical directory sandboxing and port bindings for \textsc{KidsWiki} (\texttt{runs/kidswiki\_rocket\_age4/}). Worker instances ($a_{\text{writer}}$, $a_{\text{fact}}$, $a_{\text{age}}$, $a_{\text{illustrator}}$) execute in isolated CWDs with local \texttt{weave.md} manifests: Stage~1 checkers read sibling \texttt{../writer/draft.md}, whereas Stage~1/2 deliverables (\texttt{01\_draft.md}, \texttt{02\_illustration.png}) promote to \texttt{\$\{RUN\_DIR\}} through declared port bindings.}
\label{fig:kidswiki_arch}
\end{figure*}

\paragraph{Directory Sandboxing.}
Every pipeline run lives under \texttt{runs/<context\_slug>/}, divided into stage folders (\texttt{stage01\_editorial/}, \texttt{stage02\_publishing/}) where each worker receives its own Current Working Directory (CWD):
\begin{equation}
    \text{CWD}(a) = \texttt{runs/<context\_slug>/<stage\_slug>/agents/[<group>/]<agent\_id>/}
\end{equation}
Optional organizational subdirectories (\texttt{[<group>/]}, such as per-act or per-chapter folders like \texttt{agents/act\_01/<agent\_id>/}) may sit ahead of \texttt{<agent\_id>/} to keep high-fan-out stages structured. Workers write scratch notes and intermediate files inside $\text{CWD}(a)$, whereas shared run deliverables live directly in \texttt{runs/<context\_slug>/} (\texttt{\$\{RUN\_DIR\}}) and are accessed strictly through declared port bindings. Unapproved edits to workspace-root files or sibling directories are prohibited. Figure~\ref{fig:kidswiki_arch} shows the folder layout and port bindings for \textsc{KidsWiki}.

Within each worker sandbox, \texttt{weave-agent} restricts filesystem activity so subagents do not waste context tokens crawling parent folders: broad directory scans (\texttt{dir}, \texttt{ls}, \texttt{Get-ChildItem}) outside $\text{CWD}(a)$ are prohibited, text reads are confined to $\text{CWD}(a)$ and explicitly bound inputs, binary deliverables (\texttt{*.png}, \texttt{*.mp3}) are checked for existence and byte size by MCP rather than loaded into LLM context, and status updates invoke MCP tools directly rather than through ad-hoc wrapper scripts.

\paragraph{Explicit Scoped Path Resolution and Port Aliasing.}
A deterministic path resolver binds declared file ports without opening global filesystem reads. Rather than relying on implicit multi-directory search orderings, paths use explicit scope prefixes:
\begin{itemize}
    \item \textbf{Agent CWD (unprefixed or \texttt{./}, \texttt{../}):} Paths without a variable prefix resolve against the worker's $\text{CWD}(a)$ (including sibling paths such as \texttt{../writer/script.md} or wildcard globs \texttt{../act\_*/architect/beats.md}).
    \item \textbf{Run Directory (\texttt{\$\{RUN\_DIR\}}):} Resolves directly to \texttt{runs/<context\_slug>/} for shared run-level deliverables (e.g., \texttt{\$\{RUN\_DIR\}/01\_draft.md} or \texttt{\$\{RUN\_DIR\}/02\_concept.md}).
    \item \textbf{Project Root (\texttt{\$\{PROJECT\_DIR\}}):} Resolves to the repository root for static templates, configs, and other global data.
\end{itemize}
To keep agent instructions decoupled from stage numbers, filenames, and run-root paths, input ports support aliasing of the form \texttt{<alias>: <target\_path>} (for example, \texttt{draft.md: \$\{RUN\_DIR\}/01\_draft.md}, or even an extensionless logical alias such as \texttt{draft: \$\{RUN\_DIR\}/01\_draft.md}). Inside the agent's Markdown instructions, the author refers simply to \texttt{draft.md} (or \texttt{draft}). During initialization and pre-flight verification (\texttt{weave\_verify}), the MCP server translates the alias to its concrete location on disk and informs the agent where the file lives.

\paragraph{Output Port Bindings and Completion Checks.}
Declared \texttt{outputs} in \texttt{weave.md} are optional (\texttt{outputs: []} for roles that produce no file deliverables). When outputs are declared, entries may be bare CWD filenames (such as \texttt{fact\_audit.md}) or mapped bindings \texttt{<local\_file>: <target\_path>} (such as \texttt{setup.md: \$\{RUN\_DIR\}/01\_setup.md}). When a worker calls \texttt{weave\_update} with status \texttt{completed}, the MCP server checks whether each declared output exists on disk. For mapped bindings, \texttt{verifyOutputFiles} copies \texttt{<local\_file>} from $\text{CWD}(a)$ to \texttt{<target\_path>} if the agent wrote it locally, or accepts \texttt{<target\_path>} directly if the file already exists there.

\subsection{Lifecycle Management via Status Manifests and MCP Service}
\label{sec:impl_lifecycle_mcp}

\weave{} records every lifecycle transition and port binding on disk in \texttt{weave.md}, keeping execution state inspectable outside chat transcripts. Both the run root and every worker directory hold a \texttt{weave.md} file combining a closed 7-key YAML frontmatter header (six core fields plus an optional \texttt{conversation\_id} session handle for surgical resumes) with an append-only Markdown log (Listing~\ref{lst:weave_schema}, shown for the \textsc{KidsWiki} Stage~2 \texttt{illustrator}).
\begin{lstlisting}[caption={Canonical \texttt{weave.md} state manifest schema illustrated on the \textsc{KidsWiki} \texttt{illustrator}.}, label={lst:weave_schema}]
---
type: "agent"
status: "in_progress"
instructions_file: ".agent/agents/stage2_illustrator.md"
inputs:
  - "draft.md: ${RUN_DIR}/01_draft.md"
outputs:
  - "illustration.png: ${RUN_DIR}/02_illustration.png"
updated_at: "2026-11-18T10:12:00.000Z"
---
# Illustrator Status
- 2026-11-18T10:12:00.000Z: `in_progress` - Extracting alt-text prompt from draft.md
\end{lstlisting}

\begin{figure}[t]
\centering
\begin{tikzpicture}[
    font=\normalsize,
    stateNode/.style={
        rectangle,
        rounded corners=6pt,
        draw=black,
        line width=1.1pt,
        minimum width=3.0cm,
        minimum height=1.25cm,
        align=center,
        fill=white,
        inner sep=4pt
    },
    initialState/.style={
        stateNode,
        fill=black!4
    },
    activeState/.style={
        stateNode,
        fill=white,
        line width=1.3pt
    },
    finalState/.style={
        stateNode,
        fill=black!6,
        line width=1.4pt
    },
    waitState/.style={
        stateNode,
        fill=white
    },
    errorState/.style={
        stateNode,
        dashed,
        fill=black!3,
        line width=1.1pt
    },
    arrow/.style={
        -{Stealth[length=6pt, width=4.5pt]},
        line width=1.1pt,
        draw=black
    },
    errorArrow/.style={
        -{Stealth[length=6pt, width=4.5pt]},
        line width=1.1pt,
        dashed,
        draw=black
    },
    edgeLabel/.style={
        font=\scriptsize,
        fill=white,
        inner sep=2pt,
        rounded corners=2pt
    }
]

    \node[activeState] (inprogress) {
        \textbf{in\_progress/*}\\[-1.5pt]
        {\footnotesize Active task execution}
    };

    \node[initialState, left=2.35cm of inprogress] (notstarted) {
        \textbf{initialized}\\[-1.5pt]
        {\footnotesize (or \texttt{not\_started})}
    };

    \node[finalState, right=2.35cm of inprogress] (completed) {
        \textbf{completed}\\[-1.5pt]
        {\footnotesize Outputs checked}
    };

    \node[waitState, below left=2.0cm and 0.9cm of inprogress] (waiting) {
        \textbf{waiting/*}\\[-1.5pt]
        {\footnotesize Awaiting peer input}
    };

    \node[errorState, below right=2.0cm and 0.9cm of inprogress] (failed) {
        \textbf{blocked/*}\\[-1.5pt]
        {\footnotesize Contract / tool failure}
    };

    \draw[arrow] (notstarted) -- node[edgeLabel, above=2pt] {\texttt{weave\_verify}}
                                 node[edgeLabel, below=2pt] {Pre-flight check} (inprogress);

    \draw[arrow] (inprogress) -- node[edgeLabel, above=2pt] {\texttt{weave\_update}}
                                 node[edgeLabel, below=2pt] {Output check} (completed);

    \draw[arrow] (inprogress) to[bend right=24] node[edgeLabel, left=2pt, pos=0.52] {Yield (\texttt{weave\_update})} (waiting);
    \draw[arrow] (waiting) to[bend right=24] node[edgeLabel, right=2pt, pos=0.48] {Wakeup (\texttt{weave\_verify})} (inprogress);

    \draw[errorArrow] (inprogress) to[bend left=18] node[edgeLabel, right=2pt, pos=0.5] {Fail-fast halt} (failed);
    \draw[errorArrow] (waiting) -- node[edgeLabel, below=2pt] {Missing input / timeout} (failed);

    \path (inprogress) edge [loop above, min distance=9mm, in=65, out=115, line width=1.1pt, -{Stealth[length=6pt, width=4.5pt]}]
        node[edgeLabel, above=2pt] {\texttt{weave\_update(status, progress)}} (inprogress);

\end{tikzpicture}
\caption{\weave{} worker state lifecycle across standardized status prefixes (orchestrator manifests additionally pause at \texttt{awaiting\_review/*} during human review gates). Pre-flight checks resolve declared inputs before work starts, and completion verifies declared outputs on disk.}
\label{fig:state_machine}
\end{figure}
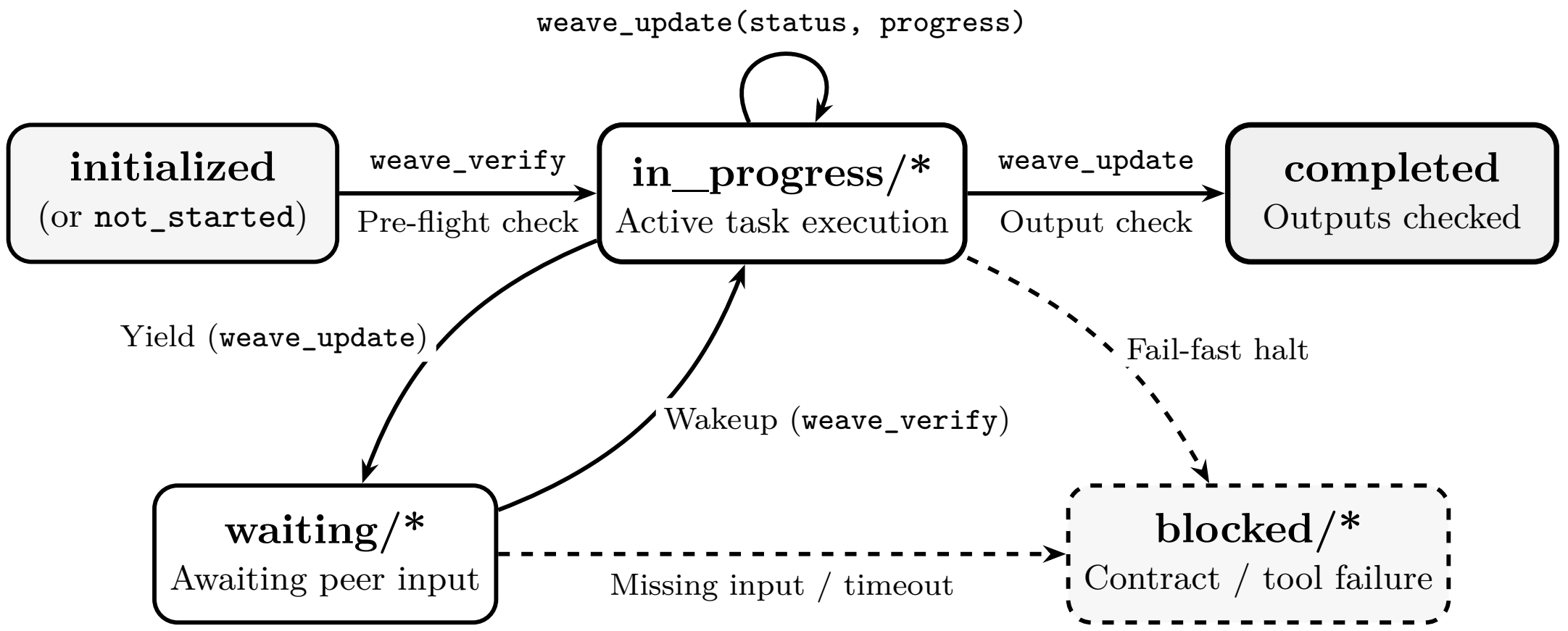

\paragraph{Native MCP Tool Suite (\mcpweave{}).}
The \mcpweave{} MCP server (namespaced by \texttt{agy} plugins as \texttt{<plugin>\_<server>}, i.e., \texttt{weavemx\_weave}) drives the state machine in Figure~\ref{fig:state_machine} through five tools:
\begin{itemize}
    \item \texttt{weave\_init}(\texttt{dir}, \texttt{type}, \texttt{status}, \texttt{instructions\_file}, \texttt{inputs}, \texttt{outputs}, \texttt{conversation\_id}): Creates a \texttt{weave.md} manifest (\texttt{type} $\in \{\texttt{"agent"}, \texttt{"orchestrator"}\}$, default \texttt{status: "not\_started"}). For workers, the Phase~1 \texttt{provisioner} sets \texttt{status: "initialized"} to mark scaffolded workspaces ready for dispatch.
    \item \texttt{weave\_verify}(\texttt{dir}, \texttt{expected\_role}): Mandatory Step~0 check called on Turn~1 of every worker run (and upon waking from \texttt{waiting/*}). It checks directory and role identity, resolves declared \texttt{inputs} (expanding \texttt{\$\{RUN\_DIR\}}, \texttt{\$\{PROJECT\_DIR\}}, aliases, and \texttt{*} globs), and partitions them into \texttt{resolved\_inputs} on disk and \texttt{pending\_inputs} (deferred peer outputs). If an input required for the worker's current task remains in \texttt{pending\_inputs}, the worker halts immediately via \texttt{weave\_update} (\texttt{blocked/*}).
    \item \texttt{weave\_update}(\texttt{dir}, \texttt{status}, \texttt{type}, \texttt{progress}, \texttt{outputs}, \texttt{conversation\_id}): Transitions lifecycle state across \texttt{[<stage>/]<state>[/<detail>]} where \texttt{<state>} is one of \texttt{not\_started}, \texttt{in\_progress}, \texttt{waiting}, \texttt{awaiting\_review}, \texttt{blocked}, or \texttt{completed} (requiring \texttt{type: "orchestrator"} when updating an orchestrator manifest), persists optional \texttt{conversation\_id} handles, and appends a timestamped log entry. When entering \texttt{completed}, it verifies declared \texttt{outputs} on disk and promotes \texttt{<local>: <target>} bindings.
    \item \texttt{weave\_read}(\texttt{dir} $|$ \texttt{dirs} $|$ \texttt{stage\_dir}, \texttt{expected\_outputs}, \texttt{expected\_status}): Lets the orchestrator confirm worker status and verify declared output files on disk for a single worker (\texttt{dir}) or across an entire wave or stage in one batch call (\texttt{dirs}, \texttt{stage\_dir}), without reading full Markdown logs or trusting chat replies.
    \item \texttt{weave\_discover\_agents}(\texttt{directory}, \texttt{filter\_status}, \texttt{limit}): Scans a stage or run directory for worker \texttt{weave.md} manifests matching \texttt{filter\_status} (default \texttt{"initialized"}), supporting paginated discovery via \texttt{limit}.
\end{itemize}
All five tools validate inputs strictly and fail fast without fallback values: invalid parameters, missing required files, or schema violations raise an immediate error or transition the node to \texttt{blocked/<reason>}, halting bad state before it propagates downstream.

\subsection{Monitoring and Out-of-Band Observability}
\label{sec:impl_monitoring_ui}

Most agent frameworks tie observability into the execution path via SDK callbacks or supervisor prompts. \weave{} separates monitoring completely through an asynchronous Node.js UI panel that periodically scans \texttt{runs/<context\_slug>/} for \texttt{weave.md} manifests using standard filesystem calls, adding zero LLM calls and zero token overhead to running agents. As illustrated on the production run in Figure~\ref{fig:tunemachine_ui}, the web panel renders a pipeline stepper showing stage progress and review states, collapsible agent status cards grouped by stage with resolved input chips and live wall-clock runtimes, single-click deliverable previews for text and media outputs, and a live activity log tailing append-only entries from \texttt{weave.md}.

\subsection{Protocol Portability vs.\ Host-Specific Bindings}
\label{sec:impl_portability}

Google Antigravity (\texttt{agy})~\cite{google2024antigravity} (app and IDE) serves as the primary reference host for the case studies, while \weave{} cleanly separates its portable coordination protocol from thin host bindings. The \texttt{weave.md} manifest schema, the Node.js \mcpweave{} MCP server, explicit scoped path resolution and port aliasing, per-agent CWD sandboxing, and the UI panel rely solely on standard POSIX/Windows filesystem calls and MCP~\cite{anthropic2024mcp}. Out of the box, \weave{} supports \texttt{agy} app, \texttt{agy} IDE, VS Code, and Cursor, alongside documented setup notes for Claude Code and Windsurf (and straightforward adaptation to other MCP-capable harnesses such as Block's Goose). Only three thin adapter points vary by host: registering subagents (such as \texttt{.agent/agents/*.md} frontmatter in \texttt{agy}), spawning and messaging sessions (such as \texttt{invoke\_subagent} and \texttt{send\_message} in \texttt{agy}, or the target harness's native subagent and MCP configuration), and opening the UI panel via host panel links or a browser view. Switching between supported or adapted harnesses changes only those thin host bindings and setup configuration, leaving agent files, stage guides, and MCP enforcement untouched.

\section{Pipeline Data-Flow Auditing and Contract Enforcement}
\label{sec:verification}
\label{sec:arch_verification}

Authoring a \weave{} pipeline across separate Markdown agent files (\texttt{.agent/agents/*.md}) and orchestrator guides (\texttt{orchestrator.md}, \texttt{orchestrator/stage*.md}) creates natural opportunities for cross-file drift: an agent might expect a runtime parameter that the orchestrator omits, write to a local filename that differs from its declared output port, or read an input file before any upstream wave produces it. To catch these errors during authoring and enforce every port boundary at runtime, \weave{} pairs three formal \emph{Data-Flow Consistency Conditions} with a two-tier auditing and enforcement architecture.

\subsection{Data-Flow Consistency Conditions}
\label{sec:arch_verification_formal}

Data-flow consistency over $G = (V, E)$ separates persistent \emph{file artifacts} $\mathcal{F}$ (resolved on disk through explicit scope prefixes and port aliases from Section~\ref{sec:impl_isolation}) from scalar \emph{runtime parameters} $\mathcal{P}$ (passed in dispatch prompts).

\paragraph{Intrinsic Agent Interfaces vs.\ Extrinsic Orchestrator Bindings.}
Each agent Markdown file is authored as a self-contained specialist that refers only to local filenames (such as \texttt{draft.md}), without hardcoding stage numbers or sibling directory paths. Formally, each worker specification $a \in \mathcal{A}$ declares an \emph{intrinsic (local) port signature}:
\begin{equation}
    \mathcal{I}_{\text{agent}}(a) = \langle F_{\text{in}}^{\text{loc}}(a), \; F_{\text{out}}^{\text{loc}}(a), \; P_{\text{in}}^{\text{loc}}(a), \; P_{\text{out}}^{\text{loc}}(a) \rangle
\end{equation}
where $F_{\text{in}}^{\text{loc}}(a)$ is the set of local file names read by $a$, $F_{\text{out}}^{\text{loc}}(a)$ is the set of local output filenames or wildcard patterns written by $a$, $P_{\text{in}}^{\text{loc}}(a)$ is the set of required input parameters (such as \texttt{topic}, \texttt{target\_age}, or slice selectors like \texttt{segment\_id}), and $P_{\text{out}}^{\text{loc}}(a)$ is the set of runtime parameters emitted by $a$ back to the orchestrator upon handoff or completion. Any of the four sets may be empty ($\emptyset$).

For each scheduled invocation $\alpha$ of $a$, the orchestrator instructions supply a matching \emph{extrinsic binding}:
\begin{equation}
    \mathcal{B}_{\text{orch}}(\alpha) = \langle \beta_{\text{in}}^{\alpha}, \; \beta_{\text{out}}^{\alpha}, \; P_{\text{in}}^{\text{orch}}(\alpha), \; P_{\text{out}}^{\text{orch}}(\alpha) \rangle
\end{equation}
defined over the orchestrator-declared local input port set $F_{\text{in}}^{\text{orch}}(\alpha)$ and local output port set $F_{\text{out}}^{\text{orch}}(\alpha)$:
\begin{itemize}
    \item $\beta_{\text{in}}^{\alpha}: F_{\text{in}}^{\text{orch}}(\alpha) \to \mathcal{F}$ maps each declared local input name $f \in F_{\text{in}}^{\text{orch}}(\alpha)$ to a concrete source path on disk: (i)~a CWD-relative or sibling path (e.g., \texttt{../writer/draft.md}, whose basename \texttt{draft.md} serves directly as the local port name), (ii)~an aliased deliverable in the run directory (e.g., $\texttt{draft.md} \mapsto \texttt{\$\{RUN\_DIR\}/01\_draft.md}$), or (iii)~a static workspace file (e.g., \texttt{\$\{PROJECT\_DIR\}/prompts/personas/*.md}).
    \item $\beta_{\text{out}}^{\alpha}: F_{\text{out}}^{\text{orch}}(\alpha) \rightharpoonup \mathcal{F}$ is a partial promotion map from declared local output names to run-directory targets (e.g., $\texttt{draft.md} \mapsto \texttt{\$\{RUN\_DIR\}/01\_draft.md}$). When an intermediate output is kept inside $\text{CWD}(\alpha)$ without run-directory promotion, no target is assigned, written $\beta_{\text{out}}^{\alpha}(f) = \bot$.
    \item $P_{\text{in}}^{\text{orch}}(\alpha) \subseteq \mathcal{P}$ is the set of scalar parameters listed explicitly on the invocation step's \texttt{Parameters:} line.
    \item $P_{\text{out}}^{\text{orch}}(\alpha) \subseteq \mathcal{P}$ is the set of emitted parameters captured by the orchestrator from $\alpha$'s handoff or completion message to populate downstream \texttt{Parameters:} dispatches.
\end{itemize}
In \textsc{KidsWiki}, \texttt{fact\_checker}, \texttt{age\_checker}, and \texttt{illustrator} all declare the exact same intrinsic file input $F_{\text{in}}^{\text{loc}} = \{\texttt{draft.md}\}$, keeping their role prompts decoupled from pipeline topology. The orchestrator binds that local port according to stage context: in Stage~1, $\beta_{\text{in}}^{\alpha_{\text{fact}}}(\texttt{draft.md}) = \texttt{../writer/draft.md}$ resolves to the sibling writer's CWD while $\beta_{\text{out}}^{\alpha_{\text{fact}}}(\texttt{fact\_audit.md}) = \bot$ keeps the critique local. In Stage~2, $\beta_{\text{in}}^{\alpha_{\text{illustrator}}}(\texttt{draft.md}) = \texttt{\$\{RUN\_DIR\}/01\_draft.md}$ binds \texttt{draft.md} to the gate-approved deliverable in \texttt{\$\{RUN\_DIR\}}.

\paragraph{Condition 1: Symmetrical Port Contract Isomorphism.}
An invocation $\alpha$ of agent $a$ is \emph{port-consistent} if and only if the local input files, local output files, input parameters, and emitted output parameters declared in the agent instructions match those declared by the orchestrator:
\begin{equation}
    F_{\text{in}}^{\text{loc}}(a) = F_{\text{in}}^{\text{orch}}(\alpha), \quad F_{\text{out}}^{\text{loc}}(a) = F_{\text{out}}^{\text{orch}}(\alpha), \quad P_{\text{in}}^{\text{loc}}(a) = P_{\text{in}}^{\text{orch}}(\alpha), \quad P_{\text{out}}^{\text{loc}}(a) = P_{\text{out}}^{\text{orch}}(\alpha)
    \label{eq:port_isomorphism}
\end{equation}
Parameters mentioned only in narrative prose or implied by folder names do not enter $P_{\text{in}}^{\text{orch}}(\alpha)$ and fail Eq.~\eqref{eq:port_isomorphism}. Across \textsc{KidsWiki}, all four roles satisfy Eq.~\eqref{eq:port_isomorphism} (with $P_{\text{out}}^{\text{loc}} = P_{\text{out}}^{\text{orch}} = \emptyset$):
\begin{itemize}
    \item \textbf{Writer ($\alpha_{\text{writer}}$):} $F_{\text{in}}^{\text{loc}} = F_{\text{in}}^{\text{orch}} = \emptyset$, $F_{\text{out}}^{\text{loc}} = F_{\text{out}}^{\text{orch}} = \{\texttt{draft.md}\}$, and $P_{\text{in}}^{\text{loc}} = P_{\text{in}}^{\text{orch}} = \{\texttt{topic}, \texttt{target\_age}\}$ (in-place revision feedback is routed via \texttt{send\_message} to the active writer session rather than as a Turn-1 file input).
    \item \textbf{Fact and Age Checkers ($\alpha_{\text{fact}}, \alpha_{\text{age}}$):} Both match on local input $\{\texttt{draft.md}\}$, local outputs $\{\texttt{fact\_audit.md}\}$ and $\{\texttt{age\_audit.md}\}$, and role-specific parameter sets $\{\texttt{topic}\}$ and $\{\texttt{target\_age}\}$. Had \texttt{stage1\_editorial.md} described age calibration only in prose while omitting \texttt{Parameters: target\_age}, $P_{\text{in}}^{\text{orch}}(\alpha_{\text{age}}) = \emptyset \neq \{\texttt{target\_age}\}$ would violate Condition~1 prior to execution.
    \item \textbf{Illustrator ($\alpha_{\text{illustrator}}$):} Matches on $F_{\text{in}}^{\text{loc}} = F_{\text{in}}^{\text{orch}} = \{\texttt{draft.md}\}$, $F_{\text{out}}^{\text{loc}} = F_{\text{out}}^{\text{orch}} = \{\texttt{illustration.png}\}$, and $P_{\text{in}}^{\text{loc}} = P_{\text{in}}^{\text{orch}} = \emptyset$ (\texttt{Parameters: none}).
\end{itemize}

\paragraph{Condition 2: Symbolic Data-Flow Execution and Causal Input Availability.}
Port consistency alone does not prevent scheduling an agent before its file or parameter inputs exist. To verify temporal ordering, symbolic waves $t = 1, 2, \dots, T$ track the available file set $\mathcal{F}_t$ (materialized across worker CWDs, \texttt{\$\{RUN\_DIR\}}, and \texttt{\$\{PROJECT\_DIR\}}) and available parameter set $\mathcal{P}_t$ at the end of wave $t$, starting from static repository files $\mathcal{F}_0 = \mathcal{F}_{\text{project}}$ and initial operator parameters $\mathcal{P}_0 = \mathcal{P}_{\text{init}}$.

An invocation $\alpha$ starts at wave $t_{\text{start}}$ and, in multi-turn handoffs, may pause in $\texttt{waiting/*}$ while peers run before resuming at wave $t_{\text{resume}} > t_{\text{start}}$. Letting $F_{\text{in}}(\alpha, \tau) \subseteq F_{\text{in}}^{\text{loc}}(a)$ denote the local file inputs read at wave $\tau \in \{t_{\text{start}}, t_{\text{resume}}\}$, causal availability requires all dispatch parameters at $t_{\text{start}}$ and every consumed file at $\tau$ to exist in the preceding wave:
\begin{equation}
    \forall \alpha \in \hat{\mathcal{A}}: \quad P_{\text{in}}^{\text{orch}}(\alpha) \subseteq \mathcal{P}_{t_{\text{start}} - 1} \quad \text{and} \quad \forall \tau \in \{t_{\text{start}}, t_{\text{resume}}\}, \; \forall f \in F_{\text{in}}(\alpha, \tau): \; \beta_{\text{in}}^{\alpha}(f) \in \mathcal{F}_{\tau - 1}
    \label{eq:causal_availability}
\end{equation}
Eq.~\eqref{eq:causal_availability} rules out same-wave read--write races while supporting both agent-emitted parameters and \emph{deferred wakeup inputs}.

In \textsc{KidsWiki}, starting from $\mathcal{F}_0 = \emptyset$ and $\mathcal{P}_0 = \{\texttt{topic}, \texttt{target\_age}\}$, all inputs are consumed at $\tau = t_{\text{start}}$ across three symbolic waves:
\begin{enumerate}
    \item \textbf{Wave $t = 1$ (\texttt{writer}):} Consumes $\{\texttt{topic}, \texttt{target\_age}\} \subseteq \mathcal{P}_0$ with $F_{\text{in}}(\alpha_{\text{writer}}, 1) = \emptyset$, satisfying Eq.~\eqref{eq:causal_availability}.
    \item \textbf{Wave $t = 2$ (parallel \texttt{fact\_checker} and \texttt{age\_checker}):} Consumes $\{\texttt{topic}\}$ and $\{\texttt{target\_age}\}$ from $\mathcal{P}_1$ and reads $\beta_{\text{in}}(\texttt{draft.md}) =$ \texttt{stage01\_editorial/agents/writer/draft.md} $\in \mathcal{F}_1$. Scheduling \texttt{writer} and \texttt{fact\_checker} concurrently in wave~$1$ would instead fail Eq.~\eqref{eq:causal_availability} since $\beta_{\text{in}}^{\alpha_{\text{fact}}}(\texttt{draft.md}) \notin \mathcal{F}_0$.
    \item \textbf{Wave $t = 3$ (Stage~2 \texttt{illustrator}, after the Stage~1 review gate):} Reads $\beta_{\text{in}}^{\alpha_{\text{illustrator}}}(\texttt{draft.md}) = \texttt{\$\{RUN\_DIR\}/01\_draft.md} \in \mathcal{F}_1 \subseteq \mathcal{F}_2$.
\end{enumerate}

\paragraph{Condition 3: Dynamic Placeholder Resolution and Monotonic State Evolution.}
When working directories or port bindings contain dynamic index placeholders (such as \texttt{writer\_chapter<CC>} for a larger writing process that requires parallel chapter-writing agents), each placeholder variable must be derivable from an upstream artifact already in $\mathcal{F}_{t-1}$. In \textsc{KidsWiki}, all four worker directories and port bindings are static singletons without index placeholders, satisfying this requirement directly. At the end of each wave $t$, the invocations $\mathcal{A}_t$ completing (or handing off parameters) in wave $t$ expand the available file and parameter sets monotonically:
\begin{align}
    \mathcal{F}_t &= \mathcal{F}_{t-1} \;\cup\; \bigcup_{\alpha \in \mathcal{A}_t} \Big( \big\{ \text{CWD}(\alpha)/f \;\big|\; f \in F_{\text{out}}^{\text{loc}}(a) \big\} \;\cup\; \big\{ \beta_{\text{out}}^{\alpha}(f) \;\big|\; f \in F_{\text{out}}^{\text{orch}}(\alpha), \; \beta_{\text{out}}^{\alpha}(f) \neq \bot \big\} \Big) \nonumber \\
    \mathcal{P}_t &= \mathcal{P}_{t-1} \;\cup\; \bigcup_{\alpha \in \mathcal{A}_t} P_{\text{out}}^{\text{orch}}(\alpha)
    \label{eq:state_evolution}
\end{align}
Manual file copying by the orchestrator outside $\beta_{\text{out}}^{\alpha}$ is prohibited. Tracing Eq.~\eqref{eq:state_evolution} across \textsc{KidsWiki} ($\mathcal{F}_0 = \emptyset$, $\mathcal{P}_t = \mathcal{P}_0 = \{\texttt{topic}, \texttt{target\_age}\}$ for all $t$ since $P_{\text{out}}^{\text{orch}} = \emptyset$) yields:
\begin{align*}
    \mathcal{F}_1 &= \big\{ \texttt{stage01\_editorial/agents/writer/draft.md}, \; \texttt{\$\{RUN\_DIR\}/01\_draft.md} \big\} \\
    \mathcal{F}_2 &= \mathcal{F}_1 \cup \big\{ \texttt{stage01\_editorial/agents/fact\_checker/fact\_audit.md}, \\
                  &\phantom{= \mathcal{F}_1 \cup \big\{\;} \texttt{stage01\_editorial/agents/age\_checker/age\_audit.md} \big\} \\
    \mathcal{F}_3 &= \mathcal{F}_2 \cup \big\{ \texttt{stage02\_publishing/agents/illustrator/illustration.png}, \\
                  &\phantom{= \mathcal{F}_2 \cup \big\{\;} \texttt{\$\{RUN\_DIR\}/02\_illustration.png} \big\}
\end{align*}
Wave~$1$ populates both the writer's local \texttt{draft.md} (consumed by sibling verifiers in wave~$2$) and the promoted \texttt{\$\{RUN\_DIR\}/01\_draft.md} (approved at the review gate and consumed by the Stage~2 illustrator in wave~$3$), whereas wave~$2$ adds only CWD critique files because $\beta_{\text{out}} = \bot$ for both verifiers.

\subsection{Two-Tier Auditing and Enforcement Architecture}
\label{sec:arch_verification_practical}

Agent instructions and orchestrator guides are written in natural-language Markdown rather than a rigid DSL, so extracting $\mathcal{I}_{\text{agent}}(a)$ from prose cannot offer compiler-grade soundness on its own. Eqs.~\eqref{eq:port_isomorphism}--\eqref{eq:state_evolution} are therefore enforced in two tiers: fast \emph{heuristic auditing} during authoring, backed by \emph{programmatic contract enforcement} at runtime.

\paragraph{Tier 1: Authoring-Time Heuristic Auditing.}
Whenever \texttt{weave-authoring} generates or edits a pipeline, it executes a structured symbolic audit across all \texttt{.agent/agents/<agent>.md} files, \texttt{orchestrator.md}, and \texttt{orchestrator/stage*.md} before any worker is launched. Stepping through symbolic waves $\mathcal{F}_t$, it prints a side-by-side contract checklist. Listing~\ref{lst:static_verification} shows the audit report produced when generating \textsc{KidsWiki}.

\begin{lstlisting}[float=!b, basicstyle=\ttfamily\footnotesize, caption={Authoring-time data-flow audit report generated for \textsc{KidsWiki}.}, label={lst:static_verification}]
### Static Pipeline Data-Flow Verification
#### Stage 1: `stage1_editorial` (`orchestrator/stage1_editorial.md`)
- [x] `writer` (`.agent/agents/stage1_writer.md`)
  - [x] Agent/Orchestrator Contract Match:
    - Agent: Reads `[none]` (Params: `topic`, `target_age`), Writes `[draft.md]`
    - Orchestrator: Inputs `[none]` (Params: `topic`, `target_age`), Outputs `[draft.md: ${RUN_DIR}/01_draft.md]`
  - [x] Input Availability (t=1): No file inputs required, receives `topic` and `target_age` via parameters
- [x] `fact_checker` (`.agent/agents/stage1_fact_checker.md`)
  - [x] Agent/Orchestrator Contract Match:
    - Agent: Reads `[draft.md]` (Params: `topic`), Writes `[fact_audit.md]`
    - Orchestrator: Inputs `[../writer/draft.md]` (Params: `topic`), Outputs `[fact_audit.md]`
  - [x] Input Availability (t=2): `../writer/draft.md` available in `writer` directory (produced at t=1)
- [x] `age_checker` (`.agent/agents/stage1_age_checker.md`)
  - [x] Agent/Orchestrator Contract Match:
    - Agent: Reads `[draft.md]` (Params: `target_age`), Writes `[age_audit.md]`
    - Orchestrator: Inputs `[../writer/draft.md]` (Params: `target_age`), Outputs `[age_audit.md]`
  - [x] Input Availability (t=2): `../writer/draft.md` available in `writer` directory (produced at t=1)
#### Stage 2: `stage2_publishing` (`orchestrator/stage2_publishing.md`)
- [x] `illustrator` (`.agent/agents/stage2_illustrator.md`)
  - [x] Agent/Orchestrator Contract Match:
    - Agent: Reads `[draft.md]` (Params: `none`), Writes `[illustration.png]`
    - Orchestrator: Inputs `[draft.md: ${RUN_DIR}/01_draft.md]` (Params: `none`), Outputs `[illustration.png: ${RUN_DIR}/02_illustration.png]`
  - [x] Input Availability (t=3): `${RUN_DIR}/01_draft.md` available in run directory (promoted in Stage 1 and gate-approved)
#### Pipeline Summary (`orchestrator.md`)
- [x] All stage outputs and final pipeline outputs are produced, auto-promoted via `## Agents` port bindings, and gate-aligned.
\end{lstlisting}

Any discrepancy (for instance, \texttt{stage1\_age\_checker} expecting \texttt{target\_age} when the execution step declares \texttt{Parameters: none}) produces an unchecked warning entry and blocks activation until fixed. Because Tier~1 relies on an LLM reading Markdown prose, it serves explicitly as a high-recall authoring check rather than a sound static type checker: an ambiguously worded prose assumption could still slip past.

\paragraph{Tier 2: Programmatic Runtime Enforcement.}
Tier~2 closes that gap at runtime using plain Node.js filesystem logic (with zero LLM calls) over the closed YAML frontmatter of \texttt{weave.md}:
\begin{enumerate}
    \item \textbf{Pre-Flight Input and Identity Check (\texttt{weave\_verify}):} On Turn~1 (and on every wakeup from \texttt{waiting/*}), the worker calls \texttt{weave\_verify}. Server code checks the working directory against the expected role, resolves \texttt{\$\{RUN\_DIR\}}, \texttt{\$\{PROJECT\_DIR\}}, CWD-relative paths, aliases, and globs (\texttt{*}), and partitions declared inputs into verified \texttt{resolved\_inputs} (\texttt{fs.existsSync}) and \texttt{pending\_inputs}. If an input required for the current task is still in \texttt{pending\_inputs}, the worker halts immediately via \texttt{weave\_update} (\texttt{blocked/*}).
    \item \textbf{Post-Flight Output Check (\texttt{weave\_update}, \texttt{weave\_read}):} When an agent transitions to \texttt{completed} with non-empty \texttt{outputs}, \texttt{weave\_update} merges output bindings via \texttt{merge\allowbreak Output\allowbreak Declarations}, checks via \texttt{verify\allowbreak Output\allowbreak Files} whether every declared output or glob match exists on disk, and copies any $\beta_{\text{out}}^{\alpha}$ promotions (such as $\texttt{draft.md} \mapsto \texttt{\$\{RUN\_DIR\}/01\_draft.md}$) into \texttt{\$\{RUN\_DIR\}} while retaining the original file inside $\text{CWD}(\alpha)$. The orchestrator then calls \texttt{weave\_read} to confirm completion and output existence on disk before advancing the stage.
    \item \textbf{Out-of-Band UI Checks (\texttt{input\_verifier.mjs}):} Alongside execution, the UI runs \texttt{checkAgentInputs} and \texttt{resolveAgentOutputs} across discovered manifests to track glob match counts, verify non-zero file sizes, and tag media types (\texttt{image}, \texttt{audio}, \texttt{video}, or text) for live inspection.
\end{enumerate}

\section{Case Studies}
\label{sec:case_studies}

To evaluate how \weave{} behaves from compact authoring examples to multi-stage production workloads, this section examines two end-to-end pipelines: \textsc{KidsWiki}, an introductory 2-stage workflow introduced in Section~\ref{sec:formal_running_example}, and \textsc{TheTuneMachine}, a 9-stage, 17-role production pipeline for long-form music history shows.

\subsection{Introductory Example: \textsc{KidsWiki}}
\label{sec:case_kidswiki}

\textsc{KidsWiki} illustrates three foundational capabilities of \weave{}: one-shot pipeline synthesis from a short natural-language brief, factual error correction through isolated writer and verifier sessions, and multimodal illustration gated by human review and in-place Markdown placeholders.

\paragraph{One-Shot Pipeline Generation.}
A central goal of \weave{} is letting authors spin up structured pipelines without writing orchestration scripts, IPC glue, or custom UI code. Invoking \texttt{weave-authoring} on a single natural-language brief produces a complete runnable setup:
\begin{quote}
\small\itshape
``Use the weave-authoring skill to create a 2-stage KidsWiki pipeline for a topic and target age (3--12, default 4). Stage 1 (Editorial): a Writer drafts the article with an \texttt{![<illustration prompt>](./02\_illustration.png)} placeholder and italicized caption. A Fact Checker and Age Checker audit accuracy and readability in parallel with an in-place revision loop, followed by a human review gate to approve the draft. Stage 2 (Publishing): an Illustrator generates the image from the alt-text prompt.''
\end{quote}
From this prompt, \weave{} emits all Markdown specifications conforming to the port model:
\begin{enumerate}
    \item \textbf{Flat Agent Definitions:} Emits four Markdown agent files in \texttt{.agent/agents/} (\texttt{stage1\_writer.md}, \texttt{stage1\_fact\_checker.md}, \texttt{stage1\_age\_checker.md}, and \texttt{stage2\_illustrator.md}), each with pinned model tiers (\texttt{model: flash}) and scoped host tools. The multi-step writer includes \texttt{\# Tasks} and \texttt{\# Lifecycle}, whereas the three single-purpose agents omit them.
    \item \textbf{Stage Guides:} Generates \texttt{orchestrator.md} alongside \texttt{stage1\_editorial.md} and \texttt{stage2\_publishing.md} in \texttt{orchestrator/}, with explicit port bindings targeting \texttt{\$\{RUN\_DIR\}/01\_draft.md} and \texttt{\$\{RUN\_DIR\}/02\_illustration.png}.
    \item \textbf{Workspace Trigger Rule:} Emits \texttt{orchestrator\_trigger.md} to bind user invocations in the workspace to the root orchestrator.
\end{enumerate}
Before any run starts, \texttt{weave-authoring} executes the Tier~1 data-flow audit (Listing~\ref{lst:static_verification} in Section~\ref{sec:arch_verification_practical}) to verify port isomorphism (Condition~1), causal input availability across waves (Condition~2), and output promotion into $\mathcal{F}_t$ (Condition~3).

\paragraph{Iterative Refinements and Hallucination Elimination.}
Single-pass generation often trades accuracy for tone when writing for young audiences. Prompted in one shot to explain ``Rockets'' to a 4-year-old, frontier models routinely produce lively but false claims:
\begin{quote}
\small\itshape
``A rocket is a super-strong space boat that flies all the way to the stars!''
\end{quote}
Friendly as it sounds, the claim is wrong: chemical rockets reach Earth orbit, the Moon, and planets inside the solar system, never distant stars. Separating drafting from verification across isolated agent sessions catches and repairs the slip cleanly. In the first pass, the Writer writes \texttt{draft.md} locally, which promotes via its output port binding to \texttt{01\_draft.md} in \texttt{\$\{RUN\_DIR\}}, after which the Fact Checker and Age Checker read \texttt{../writer/draft.md} in parallel and flag the interstellar error and vocabulary issues in \texttt{fact\_audit.md} and \texttt{age\_audit.md}. On the second pass, the orchestrator forwards the combined feedback to the existing Writer session via \texttt{send\_message} (capped at 2 revision cycles) rather than spawning a fresh writer from scratch, and the Writer revises \texttt{01\_draft.md} in place:
\begin{quote}
\small\itshape
``A rocket is a super-strong space machine that flies high above the Earth into outer space! \dots{} Rockets carry astronauts to orbit the Earth and visit the Moon.''
\end{quote}
The Fact Checker and Age Checker then re-check the revised draft, confirm accuracy and readability, and mark their manifests \texttt{completed}. Running the exact same pipeline with $\texttt{target\_age} = 11$ scales the article up to a 250-word explanation of Newton's Third Law, liquid propellants, and multi-stage orbital mechanics, with both verifiers calibrating their checks to the older reading level.

\paragraph{Human Review Gates and Multimodal Asset Synthesis.}
Generating images, speech, or video adds substantial latency and inference cost, making eager generation wasteful when upstream prose still needs factual fixes or editorial changes. \weave{} avoids wasted multimodal calls by coupling a Stage~1 review gate with an in-place Markdown placeholder contract. During Stage~1, the Writer embeds a Markdown image tag whose alt-text holds a complete visual prompt followed immediately by an italicized caption (Listing~\ref{lst:image_placeholder}).
\begin{lstlisting}[caption={In-place illustration placeholder authored by the Writer.}, label={lst:image_placeholder}]
![A friendly, chunky storybook illustration for a 4-year-old showing a bright red and white rocket blasting off from a launchpad into a blue sky curving into dark outer space, with bright orange flames shooting straight down from its engines and Earth's white clouds below.](./02_illustration.png)
*3... 2... 1... Blast off! A rocket zooms up past the clouds!*
\end{lstlisting}
Once both Stage~1 verifiers pass, the orchestrator pauses at the review gate and presents \texttt{01\_draft.md} to the operator, who can inspect both the verified prose and the embedded image prompt in the UI, approving the file as-is or requesting a quick prompt tweak without re-running fact checking. Upon approval, Stage~2 dispatches the Illustrator, which reads \texttt{\$\{RUN\_DIR\}/01\_draft.md} via its aliased input port binding (\texttt{draft.md: \$\{RUN\_DIR\}/01\_draft.md}), extracts the alt-text prompt, and writes \texttt{\$\{RUN\_DIR\}/02\_illustration.png}. Because \texttt{01\_draft.md} already links to \texttt{./02\_illustration.png} inside \texttt{\$\{RUN\_DIR\}}, the final illustrated article renders in place without altering a word of verified prose.

\subsection{Production Case Study: \textsc{TheTuneMachine}}
\label{sec:case_tunemachine}

\textsc{TheTuneMachine} is a production multi-agent workflow built on \weave{} to research, script, voice, render, and publish long-form music history episodes on YouTube. Unlike \textsc{KidsWiki}, it coordinates 17 specialized agent roles across 9 sequential stages, 5 user review gates, and dynamic act- and segment-level parallel fan-out.

\input{figures/tunemachine_dag}

\paragraph{Nine-Stage Architecture and Fan-Out Topology.}
Figure~\ref{fig:tunemachine_dag} maps the full pipeline specification declared in \texttt{orchestrator.md} and \texttt{orchestrator/stage*.md}. Stage~1 (\textit{Setup}) scopes the episode brief and persona (\texttt{01\_setup.md}, \texttt{01\_host.md}). In Stage~2 (\textit{Concept}), the \texttt{Showrunner} designs the initial arc from both Stage~1 files, dispatches \texttt{track\_queries} to 4 parallel \texttt{YouTube Sourcer} workers, and calibrates runtimes in a re-sourcing loop before emitting \texttt{02\_concept.md}. Stage~3 (\textit{Writing}) fans out one \texttt{Architect} worker per act defined by the \texttt{Showrunner} to produce act-by-act \texttt{beats.md} files, feeds them to a single \texttt{Writer} authoring \texttt{03\_script.md}, and then runs parallel \texttt{Fact Checker} instances (one per spoken host segment), an \texttt{Editor}, and a blind \texttt{Cold Reader} concurrently against the draft in a consolidated revision loop. Downstream media stages form two parallel asset streams that reconverge during video assembly and upload. Stage~4 (\textit{Staging}) produces \texttt{04\_staged\_script.md} (feeding Stage~5 \texttt{Promo}, Stage~6 \texttt{Audio}, and Stage~7 \texttt{Subtitler}) alongside \texttt{04\_transitions\_telemetry.json} (feeding Stage~6 \texttt{Audio}). Stage~5 (\textit{Promo}) runs \texttt{Promo} and \texttt{Background} (from \texttt{02\_concept.md}) in parallel to feed \texttt{Thumbnail}, while Stage~6 (\textit{Audio}) renders \texttt{06\_audio/*.mp3} and \texttt{06\_audio\_manifest.md}. Stage~7 (\textit{Video}) aligns parallel \texttt{Subtitler} shards (one per spoken host segment) against \texttt{06\_audio/*.mp3} and \texttt{04\_staged\_script.md} before \texttt{Video} muxes the audio tracks, \texttt{.srt} subtitles, \texttt{05\_promo\_background.png}, and \texttt{05\_promo.md} into \texttt{07\_video/*.mp4}. Stage~8 (\textit{YouTube}) uploads the rendered MP4s with \texttt{05\_promo.md} and \texttt{05\_thumbnail.png} and interleaves music tracks from \texttt{02\_concept.md}, and Stage~9 (\textit{Publish}) switches the episode public using \texttt{08\_youtube\_manifest.md} and \texttt{05\_promo.md}. Stages~1, 2, 3, 7, and~8 pause at explicit user review gates, whereas intermediate stages~4--6 advance automatically once Stage~3's script is approved.

\begin{figure}[p]
\centering
\includegraphics[width=0.92\linewidth,height=0.84\textheight,keepaspectratio]{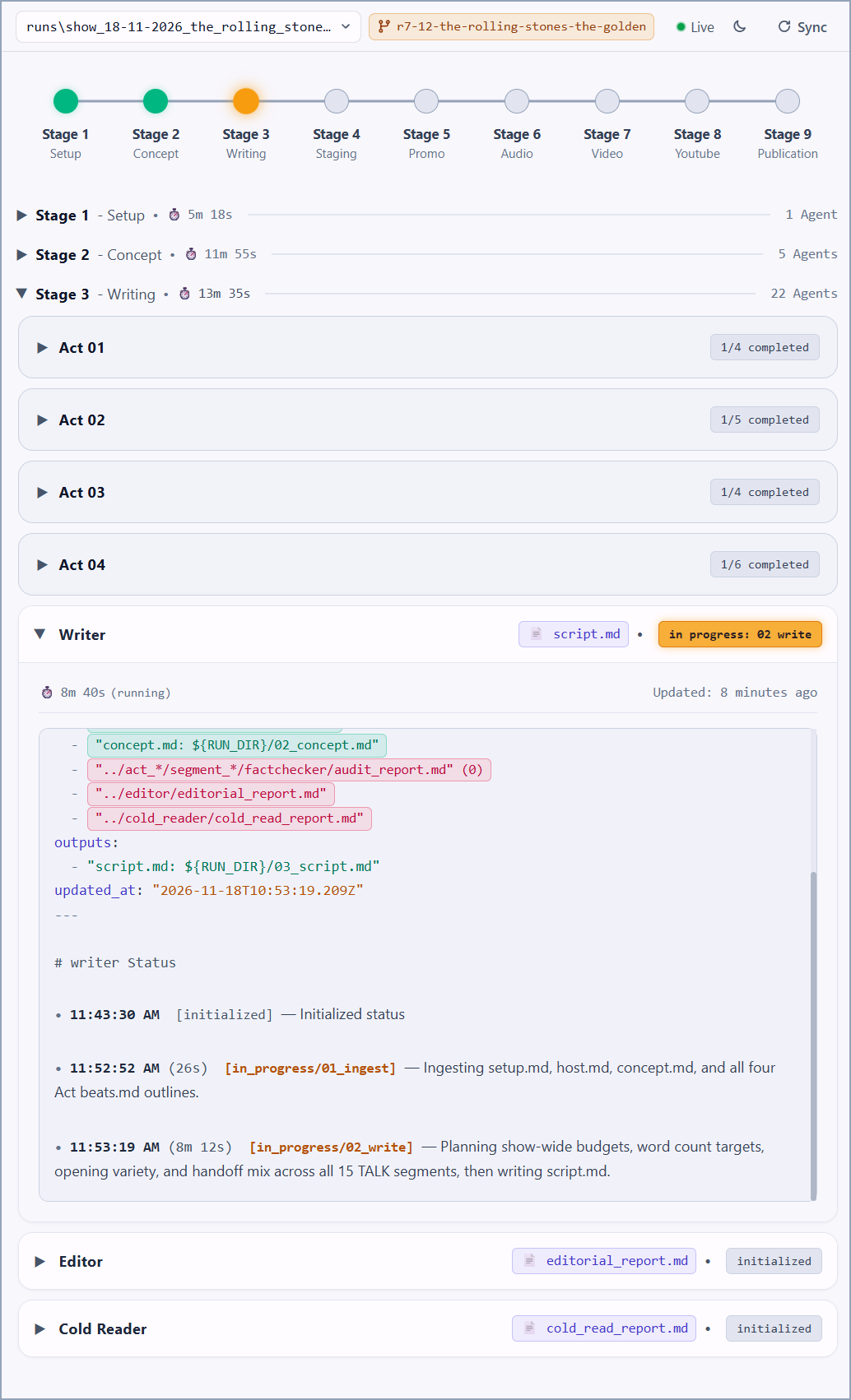}
\caption{\weave{} UI panel monitoring \textit{The Rolling Stones: The Golden Run (1968--1972)} live during Stage~3 (\textit{Writing}, $51$ worker workspaces across $9$ stages). Completed Stages~1--2 are collapsed ($1$ and $5$ agents), whereas active Stage~3 ($22$ agents) is expanded with its four act drawers (\texttt{Act~01}--\texttt{Act~04}), the \texttt{Writer} card open in \texttt{in\_progress: 02\_write} (showing resolved green vs.\ pending red input badges and timestamped task transitions), and \texttt{Editor} and \texttt{Cold Reader} queued in \texttt{initialized}.}
\label{fig:tunemachine_ui}
\end{figure}

\paragraph{Live UI Observability at Scale.}
To evaluate the finalized 17-role pipeline end to end, \textsc{TheTuneMachine} generated \textit{The Rolling Stones: The Golden Run (1968--1972)} (\texttt{show\_18-11-2026\_the\_rolling\_stones\_golden\_run}, initialized at $60$ minutes, extended to $90$ minutes via Stage~1 review feedback, and spanning $4$ acts and $15$ spoken commentary segments interleaved with $15$ music tracks). Across all $9$ stages, active execution completed in $1$~hour $23$~minutes (excluding wait times at human review gates) and materialized $51$ isolated worker \texttt{weave.md} manifests alongside the root run manifest, with every worker workspace spawned once ($51$ worker invocations plus $9$ stage-level \texttt{provisioner} invocations, $60$ \texttt{invoke\_subagent} calls in total). Figure~\ref{fig:tunemachine_ui} captures the live \weave{} UI panel during Stage~3: the 9-stage stepper highlights Stages~1--2 complete and Stage~3 active, collapsible act drawers group the completed \texttt{Architect} shards alongside the $15$ waiting \texttt{Fact Checker} shards, and the expanded \texttt{Writer} card exposes resolved vs.\ pending port bindings and append-only task logs without terminal log scraping.

\begin{figure}[p]
\centering
\begin{tikzpicture}[font=\normalsize]
\begin{axis}[
    xbar,
    bar shift=0pt,
    bar width=5.4pt,
    width=0.76\linewidth,
    height=12.0cm,
    xmin=0,
    xmax=3.05,
    xlabel={Uncached Tokens (Input + Output, Millions)},
    xlabel style={font=\footnotesize},
    xtick={0,0.5,1.0,1.5,2.0,2.5,3.0},
    xticklabel style={font=\footnotesize},
    ytick={0,1,2,3,4,5,6,7,8,9,10,11,12,13,14,15,16,17,18,19,20,21,22,23,24,25,26,27},
    yticklabels={
        {Publication (S9)},
        {\textbf{Stage Orch.\ (S9)}},
        {YouTube (S8)},
        {\textbf{Stage Orch.\ (S8)}},
        {Video (S7)},
        {Subtitler (S7, $\times 15$)},
        {\textbf{Stage Orch.\ (S7)}},
        {Audio (S6)},
        {\textbf{Stage Orch.\ (S6)}},
        {Thumbnail (S5)},
        {Background (S5)},
        {Promo (S5)},
        {\textbf{Stage Orch.\ (S5)}},
        {Staging (S4)},
        {\textbf{Stage Orch.\ (S4)}},
        {Cold Reader (S3)},
        {Editor (S3)},
        {Fact Checker (S3, $\times 15$)},
        {Writer (S3)},
        {Architect (S3, $\times 4$)},
        {\textbf{Stage Orch.\ (S3)}},
        {Showrunner (S2)},
        {YT Sourcer (S2, $\times 4$)},
        {\textbf{Stage Orch.\ (S2)}},
        {Director (S1)},
        {\textbf{Stage Orch.\ (S1)}},
        {Provisioner ($\times 9$)},
        {\textbf{Root Orchestrator}}
    },
    yticklabel style={font=\scriptsize},
    y dir=normal,
    enlarge y limits=0.025,
    xmajorgrids=true,
    grid style={black!12},
    axis lines*=left,
    axis on top=false,
    legend style={
        at={(0.98,0.89)},
        anchor=north east,
        font=\scriptsize,
        draw=black,
        fill=white,
        cells={anchor=west}
    },
    nodes near coords,
    nodes near coords align={horizontal},
    every node near coord/.append style={font=\scriptsize, anchor=west, xshift=1.5pt},
    point meta=explicit symbolic
]

\foreach \y in {1,3,6,8,12,14,20,23,25,27} {
    \edef\temp{\noexpand\fill[black!8] (axis cs:0,\y-0.46) rectangle (axis cs:3.05,\y+0.46);}%
    \temp
}

\addlegendimage{area legend, fill=black!85, draw=black}
\addlegendentry{Gemini 4 Carbon (\texttt{inherit}, S1--S5)}
\addlegendimage{area legend, fill=black!35, draw=black}
\addlegendentry{Gemini 4 Hydrogen (\texttt{inherit}, S6--S9)}
\addlegendimage{area legend, fill=white, draw=black}
\addlegendentry{Gemini Flash (\texttt{model: flash})}
\addlegendimage{area legend, fill=black!12, draw=black!45, densely dashed}
\addlegendentry{\textbf{Highlighted row:} Root / Stage Orch.}

\addplot[forget plot, fill=black!35, draw=black] coordinates {
    (0.950, 27) [\textbf{0.95M}]
    (0.037, 8)  [\textbf{0.04M}]
    (0.101, 6)  [\textbf{0.10M}]
    (0.046, 3)  [\textbf{0.05M}]
    (0.030, 1)  [\textbf{0.03M}]
};

\addplot[forget plot, fill=black!85, draw=black] coordinates {
    (0.540, 27) []
    (0.062, 25) [\textbf{0.06M}]
    (0.117, 24) [0.12M]
    (0.098, 23) [\textbf{0.10M}]
    (0.128, 21) [0.13M]
    (0.362, 20) [\textbf{0.36M}]
    (0.378, 19) [0.38M]
    (0.960, 18) [0.96M]
    (0.369, 16) [0.37M]
    (0.132, 15) [0.13M]
    (0.064, 14) [\textbf{0.06M}]
    (0.171, 13) [0.17M]
    (0.050, 12) [\textbf{0.05M}]
    (0.125, 11) [0.13M]
};

\addplot[forget plot, fill=white, draw=black] coordinates {
    (0.190, 26) [0.19M]
    (0.143, 22) [0.14M]
    (2.566, 17) [2.57M]
    (0.031, 10) [0.03M]
    (0.032, 9)  [0.03M]
    (0.133, 7)  [0.13M]
    (0.928, 5)  [0.93M]
    (0.098, 4)  [0.10M]
    (0.163, 2)  [0.16M]
    (0.056, 0)  [0.06M]
};

\end{axis}
\end{tikzpicture}
\caption{Uncached token consumption (input plus output) ordered by pipeline stage for \textit{The Rolling Stones: The Golden Run (1968--1972)} ($8.52\,\text{M}$ uncached tokens total across workers and hierarchical stage orchestration, prior to the post-release surgical edit). Shaded rows with bold labels highlight the \texttt{Root Orchestrator} ($0.95\,\text{M}$, split across $0.54\,\text{M}$ Gemini~4~Carbon in Stages~1--5 and $0.41\,\text{M}$ Gemini~4~Hydrogen in Stages~6--9) and the nine per-stage \texttt{Stage Orch.\ (S1)}--\texttt{(S9)} sub-orchestrators ($0.85\,\text{M}$ combined, peaking at $0.36\,\text{M}$ in Stage~3). All eight \texttt{model: inherit} worker roles ran on Gemini~4~Carbon in Stages~1--5, whereas high-fan-out and media roles plus \texttt{Provisioner} ($\times 9$) ran on \texttt{model: flash}.}
\label{fig:tunemachine_tokens}
\end{figure}
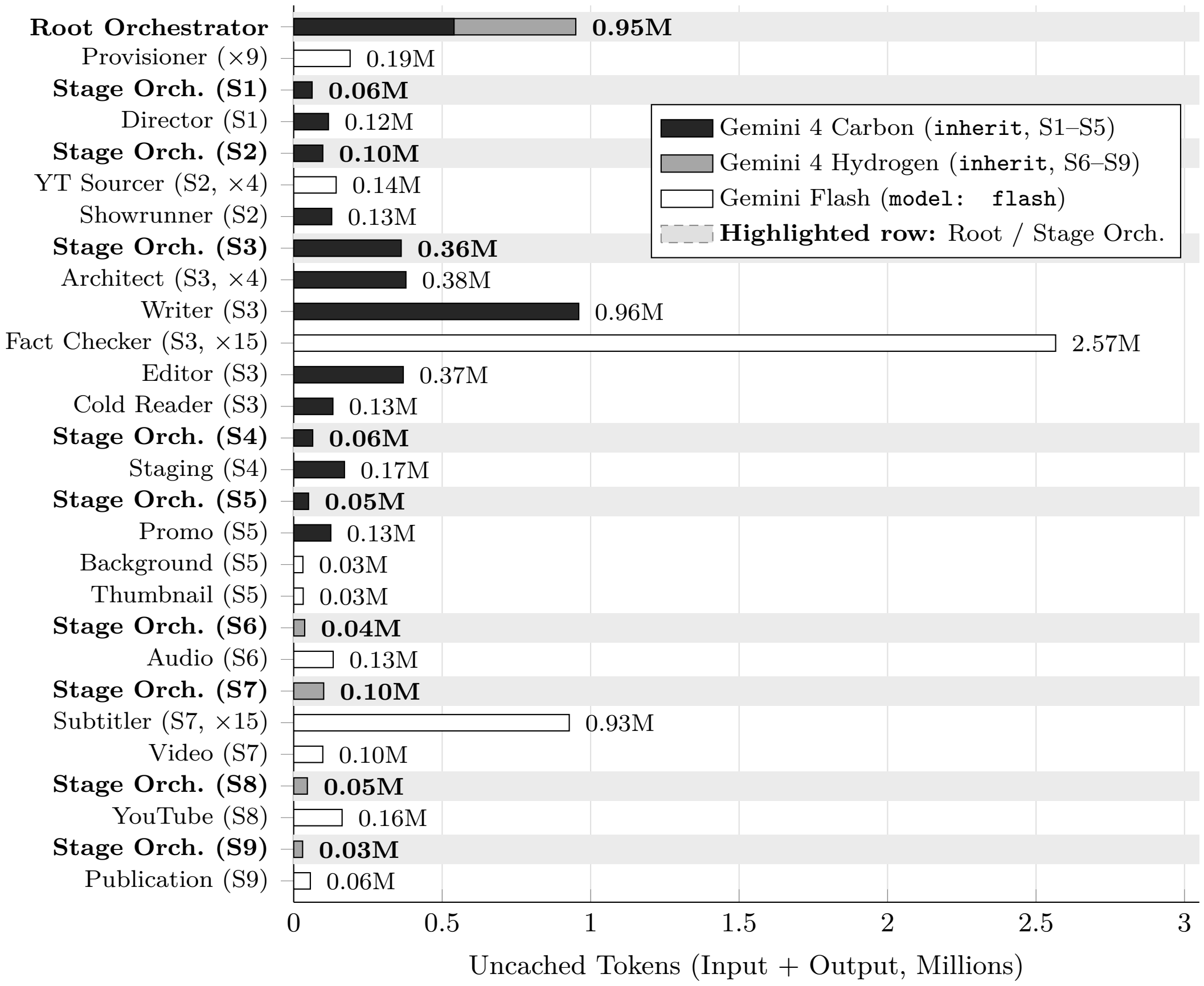

\paragraph{Per-Role Token Telemetry and Model-Tier Attribution.}
Figure~\ref{fig:tunemachine_tokens} breaks down uncached token consumption ordered by stage across the $1$~hour $23$~minute \textit{Rolling Stones} run ($8.52\,\text{M}$ uncached tokens in total prior to post-release edits). The central takeaway is that orchestration overhead remains modest relative to worker generation: isolating intra-stage dispatch, 1-line acknowledgments, and batch \texttt{weave\_read} checks inside ephemeral per-stage sub-orchestrators (\texttt{Stage Orch.\ (S1)}--\texttt{(S9)}, $0.85\,\text{M}$ combined) keeps the \texttt{Root Orchestrator} context compact ($0.95\,\text{M}$), holding combined coordination to $21.1\%$ of episode uncached tokens. Nearly four-fifths of the token budget goes directly into content creation and verification, concentrated in the $15$-way parallel \texttt{Fact Checker} loops, full-episode \texttt{Writer} drafting, and $15$-way \texttt{Subtitler} alignment.

Model-tier routing further bounds inference cost while exposing a practical avenue for future savings. Pinning high-fan-out retrieval, fact-checking, subtitle alignment, and media/CLI roles (plus \texttt{Provisioner}) to \texttt{model: flash} keeps heavy parallel waves off the frontier reasoning quota, reserving Gemini~4~Carbon for the eight creative, structural, and editorial roles in Stages~1--5. Because the case-study reference environment (\texttt{agy}) currently provides no explicit \texttt{model: carbon} selector (and \texttt{model: pro} does not map to Carbon), those Stages~1--5 roles rely on \texttt{model: inherit} from a Carbon parent session, which requires running the \texttt{Root Orchestrator} and \texttt{Stage Orch.\ (S1)}--\texttt{(S5)} on Carbon as well before switching the session to Gemini~4~Hydrogen for Stages~6--9. Supporting direct frontier model selectors on worker definitions will allow both root and stage orchestrators to run entirely on lightweight tiers, lowering coordination cost further without altering pipeline specifications.

\paragraph{Cross-Stage Surgical Asset Edits.}
In real creative production, operators rarely treat the initial public release as final: reviewing a published deliverable often surfaces a localized phrasing or timing fix upstream. In compiled state-graph frameworks, rewinding to an upstream checkpoint and modifying state replays downstream graph edges unless custom skip predicates and per-shard filters were coded ahead of time. Because \weave{} stores every stage deliverable as a plain file on disk and preserves stateful worker conversations alongside declarative \texttt{weave.md} manifests, the operator can instruct the orchestrator in natural language to patch a single segment and cascade only the affected artifacts across downstream stages.

Immediately after Stage~9 published the 30-item \textit{Rolling Stones} YouTube playlist, the operator spotted two calendar years clashing in one sentence of Segment~01 (\texttt{Act01\_Segment01}) and issued a constrained surgical instruction:
\begin{quote}
\small\itshape
``Let's do a surgical edit: `That road starts in London during the spring of 1968. Singer Mick Jagger and guitarist Keith Richards were still rattled from 1967 drug busts and courtroom trials.' \dots{} change it to \dots{} `from the previous year's drug busts and courtroom trials.' and then re-run the pipeline just for this segment and then replace the video in the playlist with the new one. Make sure not to retrigger the fact checkers / editor / cold reader for such a minor change.''
\end{quote}
Without calling \texttt{invoke\_subagent} once ($0$ new worker spawns), the orchestrator propagated the single-segment update across six stages (Stages~3, 4, 6, 7, 8, and~9) via $8$ \texttt{send\_message} dispatches to $7$ existing worker sessions:
\begin{enumerate}
    \item \textbf{Script and Staging Patch (Stages~3--4):} The orchestrator resumed \texttt{stage3\_writer} to patch \texttt{Act01\_Segment01} in \texttt{03\_script.md} while leaving all $17$ Stage~3 verifier manifests ($15$ \texttt{Fact Checker}s, \texttt{Editor}, and \texttt{Cold Reader}) untouched, then resumed \texttt{stage4\_staging} to sync \texttt{04\_staged\_script.md} and skipped Stage~5 (\texttt{Promo}, \texttt{Background}, \texttt{Thumbnail}) altogether.
    \item \textbf{Single-Segment Media Re-Synthesis (Stages~6--7):} The orchestrator resumed \texttt{stage6\_audio} to regenerate only \texttt{segment\_01\_spoken.mp3}, woke \texttt{subtitler\_s01} alone (leaving the other $14$ odd-indexed spoken shards \texttt{subtitler\_s03}--\texttt{s29} idle) to realign \texttt{segment\_01.srt}, and resumed \texttt{stage7\_video} to re-render \texttt{segment\_01\_video.mp4}, preserving the other $14$ mastered MP3, SRT, and MP4 files on disk.
    \item \textbf{Playlist Swap and Re-Publication (Stages~8--9):} Finally, the orchestrator resumed \texttt{stage8\_youtube} to upload the replacement MP4, swap it into slot~$0$ of the live YouTube playlist, and delete the superseded video, then resumed \texttt{stage9\_publication} to switch the replacement video to \texttt{public} and update \texttt{09\_publication\_manifest.md}.
\end{enumerate}
The entire 6-stage single-segment update completed in $14$~minutes ($19$ orchestrator steps and $52$ subagent steps) and consumed $0.96\,\text{M}$ uncached tokens ($0.88\,\text{M}$ input and $0.08\,\text{M}$ output, plus $10.49\,\text{M}$ cached input tokens, for $11.45\,\text{M}$ total), amounting to $11.3\%$ of the $8.52\,\text{M}$ full-episode uncached budget while retaining full append-only provenance across every modified \texttt{weave.md} manifest.

\section{Future Work and Concluding Remarks}
\label{sec:conclusion}

Declarative, file-grounded orchestration makes structured multi-agent workflows predictable and lightweight on local hardware while opening clear avenues for structural optimization and production compilation.

\subsection{Future Work}
\label{sec:future_work}

\paragraph{Low-Cost Structural and Multi-Objective Pipeline Optimization.}
A primary research direction is systematic \emph{structural pipeline optimization} over the valid specification space $\Sigma_{\text{valid}}$. Whereas single-agent prompt and instruction tuning is already extensively covered in existing literature, optimizing pipeline structure (stage boundaries, parallel fan-out width, verification-loop topology, port granularity, and per-role model tiers) remains largely open. Because a \weave{} pipeline $\sigma = \langle \mathcal{A}, \mathcal{S} \rangle$ (with extracted interfaces $\mathcal{I}_{\text{agent}}$ and orchestrator bindings $\mathcal{B}_{\text{orch}}$) exposes stage parallelization, model tiers, and port bindings in plain Markdown, it supports both \emph{automatic} exploration (in the spirit of DSPy~\cite{khattab2024dspy} and GPTSwarm~\cite{zhuge2024gptswarm}) and \emph{semi-automatic}, human-guided tuning integrated with the feedback triage workflow from Section~\ref{sec:impl_stages_authoring}. In contrast to existing automated optimization approaches that execute extensive end-to-end iterations with large token consumption, future work on \weave{} will specifically focus on \emph{low-cost, sample-efficient optimization}. Future optimization methods in \weave{} will explore the Pareto front across three competing objectives without collapsing them into a single scalar score: \textbf{run-time} (tuning parallel fan-out width, merging sequential stages where context windows allow, and overlapping deferred-wakeup work), \textbf{token usage} (per-role model down-selection, tighter input views, and early-exit rules that trim unnecessary revision passes), and \textbf{output quality} (scored by rubric evaluators, Fact-Checker pass rates, and operator preferences). Filtering structural edits through Tier~1 auditing and Tier~2 runtime checks keeps candidate pipelines inside $\Sigma_{\text{valid}}$, preserving port consistency and file provenance throughout exploration.

\paragraph{Code-Graph Compilation.}
Another planned direction is a compiler backend that lowers verified \weave{} Markdown pipelines into typed state-graph code such as LangGraph~\cite{langchain2024langgraph}, letting authors develop and run interactive workflows directly in \weave{}, while optionally compiling fixed topologies into headless, high-throughput backend services.

\subsection{Concluding Remarks}
\label{sec:conclusion_summary}

This paper presented \weave{}, a declarative orchestration layer for agent harnesses whose main contribution is the zero-code definition of structured multi-agent systems in plain Markdown. Predictable execution follows from a clear causal design: orchestrators and isolated worker agents are defined in partially predefined Markdown files, dedicated orchestrator and worker skills (\texttt{weave-orchestrator} and \texttt{weave-agent}) ensure those definitions are executed through repeatable protocols, and programmatic \mcpweave{} MCP tools enforce port boundaries and manage execution lifecycles in on-disk \texttt{weave.md} manifests. Modeling worker agents via explicit file and parameter port signatures ($a = \langle F_{\text{in}}, F_{\text{out}}, P_{\text{in}}, P_{\text{out}} \rangle$) whose cyclic specifications unwind into acyclic provenance DAGs $\hat{G} = (\hat{V}, \hat{E})$ grounds both authoring-time data-flow auditing and runtime enforcement, bridging code-heavy state graphs and unconstrained agent harnesses. Together with stateful worker sessions for natural-language surgical asset edits and a zero-token filesystem UI, these mechanisms make structured multi-agent workflows practical, inspectable, and steerable on everyday local hardware.

\bibliographystyle{plain}
\bibliography{references}

\end{document}